\documentclass[10pt,twocolumn,letterpaper]{article}

\usepackage{wacv}              

\usepackage{xcolor}         
\usepackage[table]{xcolor}
\definecolor{myblue}{RGB}{0, 102, 204}  
\definecolor{lightlightgray}{gray}{0.9}
\definecolor{positive}{HTML}{4CAF50}

\usepackage{algorithm}

\usepackage{algorithmic}
\usepackage{multirow}
\usepackage{amsthm}
\usepackage{wrapfig}
\newtheorem{definition}{Definition}
\definecolor{wacvblue}{rgb}{0.21,0.49,0.74}
\usepackage[pagebackref,breaklinks,colorlinks,allcolors=wacvblue]{hyperref}

\usepackage[accsupp]{axessibility}  

\def\wacvPaperID{2886} 
\def\confName{WACV}
\def\confYear{2026}

\title{MaxInfo: A Training-Free Key-Frame Selection Method Using Maximum Volume for Enhanced Video Understanding}

\author{Pengyi Li\textsuperscript{1, 2} \and Irina Abdullaeva\textsuperscript{1, 2, 4} \and Alexander Gambashidze\textsuperscript{1, 2} \and Andrey Kuznetsov\textsuperscript{1, 2, 4} \and Ivan Oseledets\textsuperscript{1, 3}\\
{\tt\small \{li.pengyi, abdullaeva, gambashidze, kuznetsov\}@fusionbrainlab.com} \\
{\tt\small ivan.oseledets@gmail.com}\\
\textsuperscript{1}AXXX, Russia \textsuperscript{2}FusionBrain Lab, Russia \\
\textsuperscript{3}Institute of Numerical Mathematics, Russia 
\textsuperscript{4}Innopolis University, Russia\\
}

\begin{document}
\maketitle

\begin{abstract}
Modern Video Large Language Models (VLLMs) often rely on uniform frame sampling for video understanding, but this approach frequently fails to capture critical information due to frame redundancy and variations in video content. We propose MaxInfo, the first training-free method based on the maximum volume principle, which is available in Fast and Slow versions and a Chunk-based version that selects and retains the most representative frames from a video. By maximizing the geometric volume formed by selected embeddings, MaxInfo ensures that the chosen frames cover the most informative regions of the embedding space, effectively reducing redundancy while preserving diversity. This method enhances the quality of input representations and improves long video comprehension performance across benchmarks. For instance, MaxInfo achieves a \textbf{3.28\% improvement} on LongVideoBench and a \textbf{6.4\% improvement} on EgoSchema for LLaVA-Video-7B. Moreover, MaxInfo boosts LongVideoBench performance by \textbf{3.47\%} on LLaVA-Video-72B and \textbf{3.44\%} on MiniCPM4.5. The approach is simple to implement and works with existing VLLMs without the need for additional training and very lower latency, making it a practical and effective alternative to traditional uniform sampling methods. Our code are available at \href{https://github.com/FusionBrainLab/MaxInfo.git}{https://github.com/FusionBrainLab/MaxInfo.git}
\end{abstract}


\begin{figure*}[ht]
    \centering
    \includegraphics[width=0.95\linewidth]{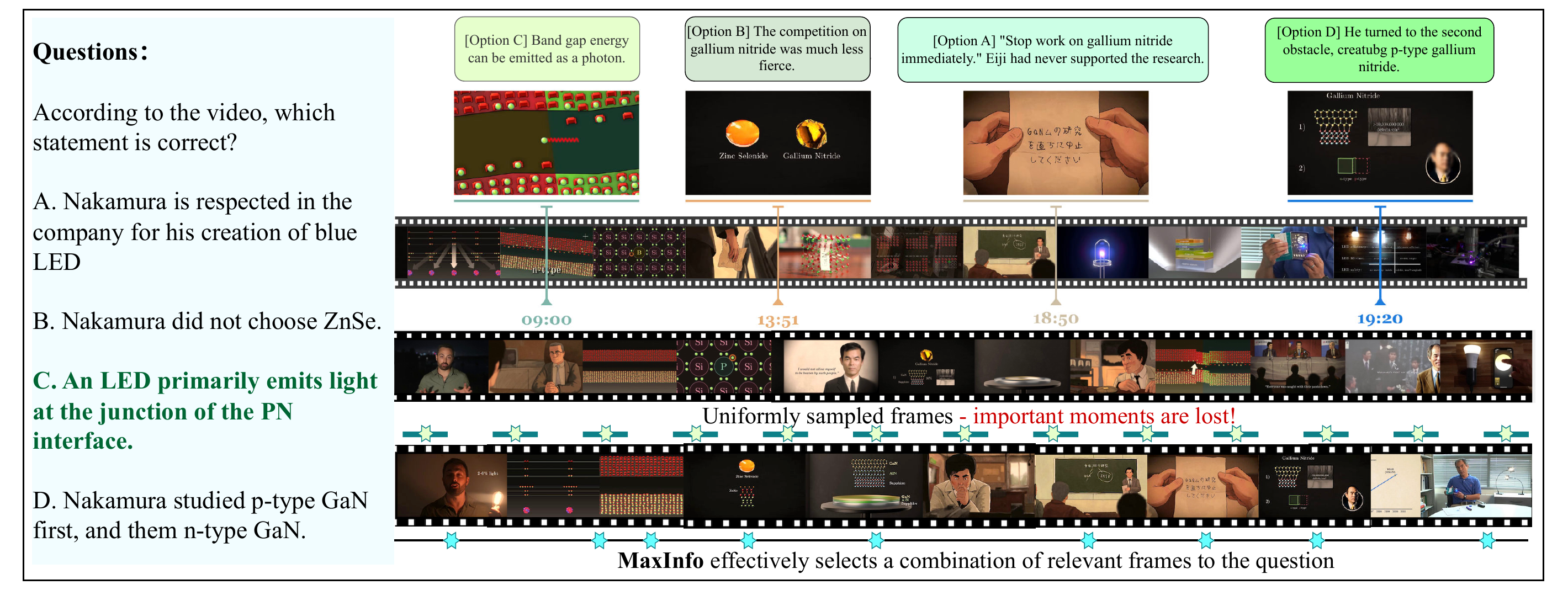}
    \caption{Reasons why the Uniform Sampling approach cannot answer the correct answer in long videos. An example of MaxInfo's sampling approach.}
    \label{video-mme-fig}
\end{figure*}

\section{Introduction}

Large language models (LLMs) such as GPT \cite{gpt-3, gpt4}, LLaMA \cite{llama2, llama3}, Qwen \cite{qwen, qwen2}, and Mistral \cite{jiang2023mistral} have revolutionized tasks like text generation, summarization, and reasoning. Recent advancements in multimodal large language models (MLLMs) \cite{llava, omnifusion} have extended these capabilities to include processing images, videos, and audio, enabling responses across diverse modalities. Video understanding, in particular, has garnered significant attention due to its complex, multi-dimensional nature and broad range of applications.

While models like LLaVA-Video \cite{llava-video}, VideoLLaMA 2 \cite{cheng2024videollama2}, MiniCPM-V 2.6 \cite{yao2024minicpmv}, and InternVL \cite{internvl} have made strides in video understanding, they struggle with long videos due to the diversity and redundancy of video content. Uniform frame sampling, a widely used approach, often fails to capture the most informative frames, leading to missing critical details and  model performance decrease. As shown in Figure~\ref{video-mme-fig} illustrate this challenge using the Video-MME \cite{video-mme} benchmark. As shown, uniform sampling can overlook key information essential to understanding the video, as frames critical to the answer may not be selected.

Existing approaches attempt to address these challenges by either increasing input sequence length or compressing video information. Models like LongVU \cite{longvu} and MovieChat \cite{song2024moviechat} compress tokens per frame, while others, such as Qwen2-VL \cite{qwen2-vl} and Gemini 1.5 Pro \cite{team2024gemini}, process longer sequences, supporting up to 32K and 10 million tokens, respectively. However, these solutions either incur substantial computational overhead or risk losing critical temporal information. Balancing the trade-off between efficiency and accuracy remains a key challenge in long video understanding.

To address this, we propose \textbf{MaxInfo} -- a training-free, plug-and-play method for dynamically selecting the most informative frames. Unlike uniform sampling, MaxInfo ensures that the input sequence maximizes information content and diversity. Our approach identifies and retains the most representative frames using the maximum volume principle on the matrix of frame embeddings and selects a subset of embeddings that span the most informative subspace. This ensures that redundant frames are removed while the retained frames span the most meaningful subspace of the video content.

Our contributions are as follows:
\begin{enumerate}[leftmargin=*]
\item A novel framework to enhance frame diversity and informativeness. MaxInfo improves upon uniform sampling by selecting the most critical frames from a video, ensuring a more meaningful representation for VLLMs.
\vspace{-0.03in}
\item An advanced scene-aware extension. We extend our framework with a scene-aware algorithm that further refines frame selection by identifying key frames within individual scenes, improving performance on tasks requiring temporal coherence.
\vspace{-0.03in}
\item Training-free and plug-and-play integration: MaxInfo requires no retraining or fine-tuning and can be seamlessly applied to any VLLM, making it a highly practical solution for long video understanding.
\vspace{-0.03in}
\end{enumerate}
\section{Related Works}

\textbf{Video Large Language Models (VLLMs).}
Video understanding has become a focal area of research, with numerous models excelling at video comprehension tasks. These tasks typically involve converting videos into image frames and inputting them into VLLMs. Existing approaches fall into two main categories: Using query-based models like Q-Former \cite{blip2} to extract critical visual features from image frames, which are then processed by VLLMs \cite{mvbench, xenos2024vllms}.
Encoding frame sequences with models such as CLIP \cite{clip}, DINO \cite{dino}, and Siglip \cite{Siglip}, and feeding the resulting embeddings into VLLMs \cite{longvu, Chat-univi, video-lavit, qwen2-vl, internvl, omnifusion, mvbench, chen2024sharegpt4video}.
While these methods emphasize visual feature extraction and text-image semantic understanding, they often rely on uniform frame sampling or similarity-based techniques, which can overlook critical information, especially in long videos with diverse content.

\textbf{Long Video Understanding.}
Understanding long videos poses significant challenges, primarily due to the need to balance computational efficiency with preserving critical temporal and contextual information. To address this, various strategies have been proposed, Reducing sequence length: Methods like Video-LaVIT \cite{video-lavit} and LongVU \cite{longvu} use cosine similarity or clustering to filter redundant frames, while MovieChat \cite{song2024moviechat} applies similarity thresholds for frame selection.
Token compression: SlowFast-LLaVA \cite{slowfast} compresses visual tokens, and Chat-UniVi \cite{Chat-univi} extracts key event tokens to reduce redundancy.
Extended input lengths: Models such as Qwen2-VL \cite{qwen2-vl} and Gemini 1.5 Pro \cite{team2024gemini} handle extended token lengths to process long videos, albeit with high computational costs.

In addition, numerous keyframe extraction algorithms have been proposed, such as LongVA \cite{zhang2024long}, Frame-Voyager \cite{yu2024frame}, AKS \cite{tang2025adaptive}, VideoTree \cite{videotree}, M-LLM \cite{hu2025m}, BOLT \cite{liu2025bolt}, VSLS \cite{guo2025logic}, AdaReTaKe \cite{wang2025adaretake}, Q-Frame \cite{zhang2025q}, GenS \cite{yao2025generative}, and ViLaMP \cite{cheng2025scaling}, all of which have demonstrated remarkable performance across various long video understanding tasks.

Despite these advancements, current methods often depend on arbitrary thresholds, fixed compression schemes, or uniform sampling, which may fail to capture the diverse and critical content of long videos effectively.

\textbf{Information Maximization Techniques.}
Information maximization is widely used for feature selection and dimensionality reduction. Methods such as mRMR \cite{peng2005feature} and MMD \cite{MMD} improve model performance by selecting features with high relevance and low redundancy, while MOI \cite{MOI} focuses on the most informative feature subsets to enhance classification. The maximum volume (MaxVol) algorithm \cite{goreinov2010find, Rectmax, sozykin2022ttopt} selects linearly independent rows of a matrix to cover the most informative subspace.

In this paper, we extends the MaxVol principle to video understanding and proposes a keyframe extraction framework tailored for VLLMs. Unlike existing methods, our approach dynamically selects diverse and representative frames. By maximizing the geometric volume of frame embeddings, MaxInfo ensures that the selected frames are both informative and diverse, and, combined with a scene-aware algorithm, enables fine-grained selection of keyframes for each video segment, providing an efficient, training-free solution for long video understanding.
\section{Method} \label{method-section}
\begin{figure}[htb!]
    \centering
    \centerline{\includegraphics[width=1\linewidth]{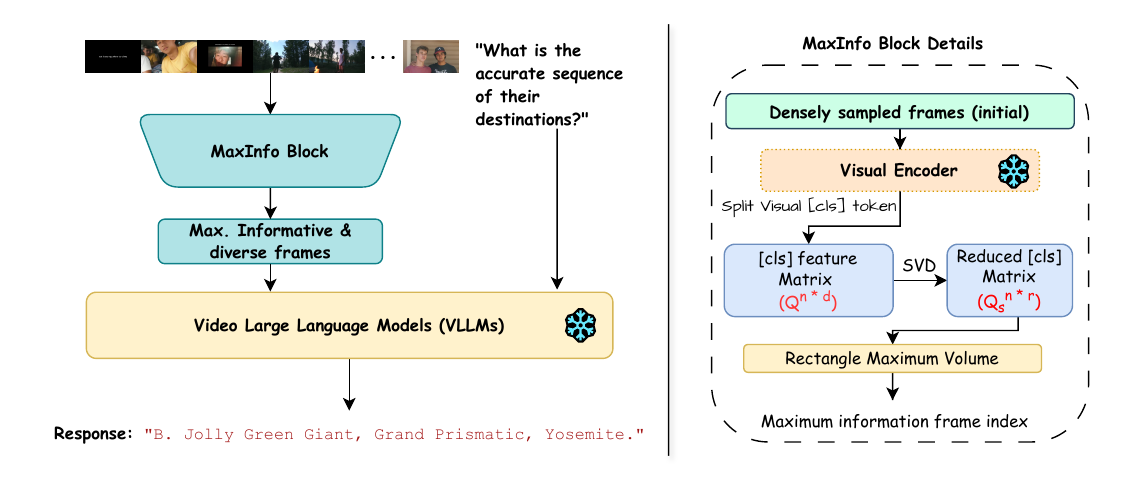}}
    \caption{\textbf{Overview of the MaxInfo Block integrated into a VLLM.} We extract the most informative frames via the MaxInfo Block and then perform inference on the resulting subset of frames.}
    \label{pipeline}
\end{figure}

We propose the \textbf{MaxInfo Block}, a plug-and-play, training-free module designed for long-video understanding tasks. It ensures both \textit{diversity} of frames and \textit{comprehensive} semantic coverage from a video by selecting only the most informative frames. As illustrated in Figure~\ref{pipeline}, the MaxInfo Block can be easily integrated into any VLLMs, enhancing the quality and diversity of visual information fed into the model. 

\subsection{Overview}
Given a video, we uniformly sample $n$ frames. For example, sampling at 1~fps reduces the risk of losing important content, but still may yield a large number of frames, many of which could be redundant. This both increases computational cost and does not guarantee capturing the \textit{most} informative or diverse frames. Hence, we seek a small, representative subset of frames.

Let the sampled frames be
$\mathbf{I} \;=\; \{\, i_1,\; i_2,\;\ldots,\; i_n \}.$
We extract each frame’s visual representation via a CLIP-based ViT \cite{clip} and retain only the [CLS] token. Stacking these tokens yields
\begin{equation}
\mathbf{Q} = \begin{bmatrix} q_1 & q_2 & \cdots & q_n \end{bmatrix}^\top
\end{equation}
where $q_n\in\mathbb{R}^d$ is the flattened [CLS] feature from the $n$-th frame.

\subsection{Dimensionality Reduction}
Handling all $n\times d$ features may still be computationally expensive when $n$ and $d$ are large. To mitigate this, we perform a truncated SVD on $\mathbf{Q}$:
\begin{equation}
    \mathbf{Q} \;=\; U\,\Sigma\,V^T \quad\rightarrow\quad \mathbf{Q}_s \;=\; U_{(:,1:s)},
\end{equation}
where $U\in \mathbb{R}^{n\times n}$, $\Sigma\in \mathbb{R}^{n\times d}$, and $V\in\mathbb{R}^{d\times d}$. By retaining the first $s$ singular vectors (the top $s$ columns of $U$), we obtain:
\begin{equation}
    \mathbf{Q}_s\;\in\;\mathbb{R}^{n\times s},
\end{equation}
which captures the principal visual variation among frames while drastically reducing dimensionality.

\subsection{Rectangular MaxVol Frame Selection}
On the next step we identify the “most informative” subset of rows of $\mathbf{Q}_s$, i.e., a set of frames which corresponding rows span the overall distribution of frames. To do this, we use the \emph{rectangular MaxVol} algorithm \cite{Rectmax} to evaluate a submatrix of maximal volume in $\mathbf{Q}_s$.

For a rectangular matrix $\mathbf{A}\in\mathbb{R}^{p\times q}$, the \emph{rect-volume} can be defined (up to transformations) as
\begin{equation}
    \text{rect-vol}(\mathbf{A}) \;=\;\sqrt{\det(\mathbf{A}\,\mathbf{A}^T)}
\end{equation}
Maximizing $\text{rect-vol}(\mathbf{A})$ with respect to the selection of rows corresponds to identifying the subset of frames that best preserves the variation in the data. We denote the selected row indices as
\begin{equation}
    \mathbf{r} \;=\; \arg\max_\mathbf{r}\,\text{rect-vol}\bigl(\mathbf{Q}_s(\mathbf{r},:)\bigr)
\end{equation}
where indices $\mathbf{r}$ then specify the frames $i_n$ we deem most representative.

\paragraph{Resulting Frame Subset.}
Let $r=|\mathbf{r}|$ be the number of selected frames. The final submatrix
\begin{equation}
    \mathbf{S} \;=\; \mathbf{Q}(\mathbf{r},:)
\end{equation}
is an $r\times n$ matrix containing the \textbf{diverse, high-information} frames. We feed only these $r$ frames into the downstream VLLM:
\begin{equation}
    \mathbf{A}_{\text{MaxInfo}} \;=\; \text{VLLM}\bigl(\text{Instruction},\;\mathbf{S},\;\text{Questions}\bigr),
\end{equation}
as opposed to using all $n$ frames (which might be computationally prohibitive or redundant):
\begin{equation}
    \mathbf{A}_{\text{Init}} \;=\; \text{VLLM}\bigl(\text{Instruction},\;\mathbf{Q},\;\text{Questions}\bigr).
\end{equation}

Given that the value of r is much smaller than n (i.e., $r\ll n$) in most application scenarios, the design significantly improves the inference efficiency while effectively avoiding the loss of critical visual context information. We innovatively propose the MaxInfo algorithm framework scene-aware MaxInfo and fast and slow versions of the implementation scheme. Experimental results show that both exhibit significant enhancements. 

\subsection{Fast and Slow Version}

Given the differing objectives of the fast and slow versions of MaxInfo, we conducted a comparison to determine the most suitable option for each experimental setup. Additionally, we compared the two variants of MaxInfo, each designed to address different computational constraints:

\textbf{Fast Version.}  
MaxInfo is applied directly to the same number of frames as the original uniform sampling.  
    For instance, if the base model uses $n$ frames, we retain $n$ frames and rely on MaxInfo to identify the most informative subset. This incurs minimal computational overhead and provides a quick improvement without altering the model’s default settings.

\textbf{Slow Version.}  
A larger pool of frames ($N \gg n$) is initially sampled to ensure extensive coverage. The MaxInfo Block is then applied to select the most diverse frames. If the resulting set $x$ exceeds the model’s maximum input limit $n$, we uniformly downsample $x \to n$. This approach offers potentially higher gains by starting with more frames, albeit at the cost of additional embedding computations.

\subsection{Chunk-Based MaxInfo}

Videos often contain multiple scenes with visually similar frames, making global frame selection suboptimal. Applied across the entire video, MaxInfo may also discard important frames due to spurious embedding similarities between different scenes.  

To address this, we propose \textbf{Chunk-Based MaxInfo}, a simple yet effective modification. We uniformly divide the video into $M$ equal-sized chunks and apply MaxInfo independently within each chunk. This ensures that every segment is adequately represented while keeping the procedure computationally efficient.

Formally, given $n$ uniformly sampled frames, we split them into $M$ contiguous chunks:
\begin{equation}
    \mathbf{I} = \bigcup_{i=1}^{M} \mathbf{I}^{(i)}, \quad \mathbf{I}^{(i)} = \{i_j \mid j \in \text{chunk } i\}.
\end{equation}
For each chunk, we extract CLIP embeddings, apply SVD for dimensionality reduction, and run MaxVol to select the most representative frames. In our experiments, we set $M=32$ for simplicity, though any choice of $M$ is possible and can be tuned to balance representation quality and computational cost.  

This approach is deliberately simple but demonstrates that even Chunk-Based scene segmentation can further enhance MaxInfo’s effectiveness. It highlights the potential for more refined scene-aware selection in future work.






\subsection{Summary}

Our pipeline can be summarized with an \cref{MaxInfo}. It is a training-free and plug-and-play algorithm that can be integrated into any model of VLLMs.

\begin{algorithm}[hpt!]
    \caption{MaxInfo Block: SVD + MaxVol for Keyframe Selection}
    \label{MaxInfo}
\begin{algorithmic}[1]
    \STATE \textbf{Input:} A set of $n$ frames \(\mathbf{I} = \{i_1, i_2, \dots, i_n\}\)
    \STATE \textbf{Embedding:} Convert each frame \(i_j\) into a [CLS] embedding:
    \[q_n = \mathrm{flatten}\bigl(\mathrm{clip}(i_n)\bigr), \mathbf{Q} = \begin{bmatrix} q_1 & q_2 & \cdots & q_n \end{bmatrix}^\top \]
    \STATE \textbf{SVD Reduction:} Perform truncated SVD on \(\mathbf{Q}\): 
    \[\mathbf{Q} \approx \mathbf{U}_r \, \mathbf{\Sigma}_r \, \mathbf{V}_r^\top \quad \rightarrow \quad \mathbf{Q_s} = \mathbf{U}_r \in \mathbb{R}^{n \times r}.\]
    \STATE \textbf{MaxVol Selection:} Run \(\text{rect\_maxvol}(\mathbf{Q_s}, \mathrm{tol})\) to find pivot indices:
    \[\mathrm{piv} = \text{rect\_maxvol}(\mathbf{Q_s}, \mathrm{Tol}),\]
    identifying rows (frames) that span the reduced embedding space.
    \STATE \textbf{Output:} Indices \(\mathrm{piv}\) of the most informative keyframes.
\end{algorithmic}
\end{algorithm}

\section{Experiments} \label{experiment}

\textbf{Overall.} In order to evaluate the contribution of MaxInfo to video understanding, we employed widely-used video understanding benchmarks, covering short-video tasks (such as MVBench \cite{mvbench}) and medium-to-long video tasks (such as EgoSchema \cite{egoschema}, Video-MME \cite{video-mme}, and LongVideoBench \cite{longvu}). For complete fairness, we only compared improved versions of the model against itself without MaxVol with freezing generation parameters, seed and prompt. 


\subsection{Main Results} \label{main-results}

\textbf{Overall Performance.}  
Table~\ref{proof-maxinfo2} present the performance gains achieved by integrating MaxInfo into existing InternVL2 \cite{internvl}, Qwen2-VL \cite{qwen2-vl} and LLaVA-Video \cite{llava-video} models. Experimental results show that MaxInfo Block exhibits significant performance improvements on a number of models. In particular, in the LLaVA-Video-7B and Qwen-VL-2B models, the introduction of MaxInfo Block improves the accuracy by 0.9\%/1.7\%, 6.4\%, 3.3\% and 1.4\%/1.2\%, 2.3\%, 1.5\% in VideoMME \cite{video-mme}, EgoSchema \cite{egoschema}, and LongVideoBench \cite{longvideobench}, respectively, which is significantly better than the versions without the MaxInfo block. This improvement not only validates the effectiveness of MaxInfo Block, but also provides new ideas for future research on video understanding tasks. 

\begin{table*}[hbt!]
\caption{Comparison of VLLM with and without MaxInfo on multiple benchmarks, where wo. sub. is without subtitles and with w. sub. subtitles.}
\label{proof-maxinfo2}
\centering
\scriptsize 
\setlength{\tabcolsep}{14pt}
\scalebox{1}{ 
\begin{tabular}{lccccc}
\toprule
Model & Size & VideoMME (wo/w-subs) & Egoshcema & LongVideoBench \\
\midrule
LLaVA-Video \cite{llava-video} & 7B & 63.3/69.7$_{(64)}$ & 57.3$_{(64)}$ & 58.2$_{(64)}$ \\
\textbf{+ MaxInfo} & 7B & 64.2/71.4$_{64 \rightarrow (6, 64)}$ & \textbf{63.7}$_{128 \rightarrow (12, 64)}$ & \textbf{61.5}$_{128 \rightarrow (1, 64)}$ \\
\rowcolor{lightlightgray} $\triangle$ & & {\scriptsize \color{positive}+0.9\%/+1.7\%} & {\scriptsize \color{positive}+6.4\%} & {\scriptsize \color{positive}+3.3\%} \\
\midrule
LLaVA-Video \cite{llava-video} & 72B & \textbf{70.5}/76.9$_{(64)}$ & 65.6$_{(64)}$ & 61.9$_{(64)}$ \\
\textbf{+ MaxInfo} & 72B & 70.9/\textbf{77.6}$_{64 \rightarrow (6, 64)}$ & \textbf{69.4}$_{128 \rightarrow (12, 64)}$ & \textbf{64.9}$_{128 \rightarrow (1, 64)}$ \\
\rowcolor{lightlightgray} $\triangle$ & & {\scriptsize \color{positive}0.4\%/\color{positive}+0.7\%} & {\scriptsize \color{positive}+3.8\%} & {\scriptsize \color{positive}+3\%} \\
\midrule
Qwen2-VL \cite{qwen2-vl} & 2B & 55.6/60.4$_{(786)}$ & 54.9$_{(180)}$ & 47.3$_{(256)}$ \\
\textbf{+ MaxInfo} & 2B & \textbf{57.0/61.6}$_{256 \rightarrow (4, 254)}$ & \textbf{57.2}$_{180 \rightarrow (12, 180)}$ & \textbf{48.8}$_{256 \rightarrow (1, 224)}$ \\
\rowcolor{lightlightgray} $\triangle$ & & {\scriptsize \color{positive}+1.4\%/+1.2\%} & {\scriptsize \color{positive}+2.3\%} & {\scriptsize \color{positive}+1.5\%} \\
\midrule
Qwen2-VL \cite{qwen2-vl} & 7B & 63.3/69.0$_{(768)}$ & \textbf{66.7}$_{(180)}$ & 53.7$_{(256)}$ \\
\textbf{+ MaxInfo} & 7B & 62.1/70.0$_{256 \rightarrow (4, 254)}$ & 64.3$_{180 \rightarrow (12, 180)}$ & \textbf{55.7}$_{256 \rightarrow (1, 224)}$ \\
\rowcolor{lightlightgray} $\triangle$ & & {\scriptsize \color{red}-1.2\%/\color{positive}+1.0\%} & {\scriptsize \color{red}-2.4\%} & {\scriptsize \color{positive}+2.0\%} \\
\midrule
InternVL2 \cite{internvl} & 1B & 43.0$_{(16)}$ & 34.0$_{(128)}$ & 43.4$_{(128)}$ \\
\textbf{+ MaxInfo} & 1B & 43.6$_{128 \rightarrow (1, 16)}$  & \textbf{34.1}$_{128 \rightarrow (1, 32)}$ & \textbf{43.9}$_{128 \rightarrow (1, 32)}$ \\
\rowcolor{lightlightgray} $\triangle$ & &  {\scriptsize \color{positive}+0.6\%} & {\scriptsize \color{positive}+0.1\%} & {\scriptsize \color{positive}+0.5\%} \\
\midrule
InternVL2 \cite{internvl} & 2B &  45.4$_{(16)}$ & 46.1$_{(128)}$ & 47.0$_{(128)}$ \\
\textbf{+ MaxInfo} & 2B &  45.0$_{128 \rightarrow (1, 16)}$ & \textbf{46.3}$_{128 \rightarrow (1, 32)}$ & \textbf{47.3}$_{128 \rightarrow (1, 32)}$ \\
\rowcolor{lightlightgray} $\triangle$ & &  {\scriptsize \color{red}-0.4\%} & {\scriptsize \color{positive}+0.2\%} & {\scriptsize \color{positive}+0.3\%} \\




\bottomrule
\end{tabular}
}
\end{table*}

Although our results are slightly lower than the baseline on some models, we use significantly fewer frames than the baseline configuration. 


\textbf{Chunck Based MaxVol.} Table~\ref{mlvu_aware} evaluates MaxInfo and Chunk-Based Scene-Awareness on the MLVU benchmark, showing consistent gains over uniform sampling. This approach is simple and entirely training-free, highlighting the potential of even basic scene-awareness. These results suggest that more advanced scene segmentation techniques could yield further improvements, making scene-awareness a promising direction for video understanding.

\begin{table}[hbt!]
  \centering
  \caption{Performance comparison on the MLVU benchmark between original Qwen2-VL and its variants with MaxVol and Chunk-Based. Best results in \textbf{bold}, second-best \underline{underlined}.}
  \label{mlvu_aware}
  \begin{small}
  \scalebox{0.90}{
  \begin{tabular}{lcc}
    \toprule
    Model & Size & Acc. \\
    \midrule
    Qwen2-VL~\cite{qwen2-vl} & 2B & 52.18\\
    \textbf{+ MaxVol} & 2B & \underline{52.37} \\
    \textbf{+ Chunk-Based} & 2B & \textbf{52.69} \\
    \midrule
    Qwen2-VL~\cite{qwen2-vl} & 7B & 64.32 \\
    \textbf{+ MaxVol} & 7B & \underline{64.59} \\
    \textbf{+ Chunk-Based} & 7B & \textbf{64.82} \\
    \bottomrule
  \end{tabular}}
  \end{small}
\end{table} 

\textbf{Comparison Baseline.} As shown in Table~\ref{same-size-sota}, our proposed MaxInfo achieves strong competitiveness among training-free keyframe extraction methods and further demonstrates comparable or superior performance relative to existing state-of-the-art approaches. 

Here, \textbf{*} indicates that our method does not require a predefined number of frames, but dynamically selects key frames based on the model’s initial inference frame count and the video’s information content. For example, on Video-MME and LongVideoBench, Qwen-MaxInfo processes an average of 180 and 170 frames, respectively, compared to 768 and 256 frames in the base Qwen model. Similarly, LLaVA-Video-MaxInfo uses 64 frames on both datasets, while the base LLaVA-Video model uses 58 and 51 frames, respectively.

\begin{table*}[htb!]
\caption{This table compares the performance of VLM with current state-of-the-art (SOTA) open-source and proprietary models, as well as key-frame selection methods, on video benchmark key-frame selection tasks. First place: \textbf{bold}, second place: \underline{underline}, third place: \textit{italic}.}
\label{same-size-sota}
\centering
\small
\setlength{\tabcolsep}{6pt}
\resizebox{0.98\textwidth}{!}{
\begin{tabular}{lcccccccc}
\toprule
\multicolumn{1}{c}{\multirow{3}{*}{\textbf{Model}}} & 
\multicolumn{1}{c}{\multirow{3}{*}{\textbf{LLM size}}} & 
\multicolumn{1}{c}{\multirow{3}{*}{\textbf{\#Frames}}} & 
\multicolumn{4}{c}{\textbf{VideoMME (wo sub.)}} & 
\multicolumn{1}{c}{\multirow{3}{*}{\textbf{EgoSchema}}} & 
\multicolumn{1}{c}{\multirow{3}{*}{\textbf{LongVideoBench (val.)}}} \\
\cmidrule(lr){4-7}
& & & Short & Medium & Long & Overall & & \\
& & & {\scriptsize \textit{1.3min}} & {\scriptsize \textit{9min}} & {\scriptsize \textit{41min}} & {\scriptsize \textit{17min}} & {\scriptsize \textit{3min}} & {\scriptsize \textit{12min}} \\
\midrule
\multicolumn{9}{c}{\textbf{Proprietary Models}} \\ 
\midrule
GPT4-o \cite{openai2024gpt4o} & - & 1fps & \textit{77.1} & 62.1 & 59.2 & 66.2 & - & \textbf{66.7} \\
Gemini-1.5-Pro \cite{team2024gemini} & - & 1fps & \textbf{82.3} & \textbf{75.3} & \textbf{67.5} & \textbf{75.7} & - & 64.0  \\
\midrule
\multicolumn{9}{c}{\textbf{Open Source Models}} \\ 
\midrule
VideoChat2 \cite{mvbench} & 7B & 16 & 48.3 & 37.0 & 33.2 & 39.5 & - & - \\
ShareGPT4Video \cite{chen2024sharegpt4video} & 8B & 16 & - & - & 37.9 & 43.6 & - & - \\
VideoLLaMA2 \cite{cheng2024videollama} & 7B & 32 & 56.0 & 45.4 & 42.1 & 47.9 & - & - \\
LongVILA \cite{xue2024longvila} & 8B & 128 & 60.2 & 48.2 & 38.8 & 49.2 & - & - \\
 & 8B & 256 & 61.8 & 49.7 & 39.7 & 50.5 & - & - \\
Qwen2-VL \cite{qwen2} & 7B & 8 & 65.0 & 50.7 & 45.3 & 53.7 & 53.5 & - \\
\midrule
\multicolumn{9}{c}{\textbf{Key-frames Selection Methods}} \\
\midrule
LongVU \cite{longvu} & 7B & 1fps & 64.7 & 58.2 & \textit{59.5} & 60.6 & \underline{67.6} & - \\
\multirow{4}{*}{LongVA~\protect\cite{zhang2024long}} & \multirow{4}{*}{7B} & 16 & 59.0 & 46.6 & 43.6 & 49.7 & - & - \\
 &  & 64 & 61.4 & 50.9 & 45.0 & 52.4 & - & - \\
 &  & 128 & 61.1 & 50.4 & 46.2 & 52.6 & - & - \\
 &  & 384 & 60.3 & 48.9 & 46.1 & 51.8 & - & - \\
Chat-Univi-v1.5 \cite{Chat-univi} & 7B & 64 & 51.2 & 44.6 & 41.8 & 45.9 & - & - \\
AKS (LLaVA-Video) \cite{tang2025adaptive} & 7B & 64 & - & - & - & 65.3 & - & 62.7 \\
M-LLM (Qwen2-VL) \cite{hu2025m} & 7B & - & 69.6 & 54.1 & 51.9 & 58.7 & \textit{65.9} & - \\
BOLT (LLaVA-OneVision) \cite{liu2025bolt} & 7B & 32 & 70.1 & 60.0 & 49.6 & 59.9 & 64.0 & 59.6 \\
AdaReTaKe (Qwen2-VL) \cite{wang2025adaretake} & 7B & - & - & - & 56.4 & 64.2 & - & 57.2 \\
AdaReTaKe (LLaVA-Video) \cite{wang2025adaretake} & 7B & - & - & - & 53.9 & 64.0 & - & 59.6 \\
GenS (Qwen2-VL \cite{yao2025generative}) & 7B & 50 & - & - & - & - & - & 58.7 \\
ViLaMP \cite{cheng2025scaling} & 7B & 1fps & - & - & 58.4 & \textit{67.7} & - & 60.2 \\
Frame-Voyager \cite{yu2024frame} & 8B & 128 (16) & 67.3 & 56.3 & 48.9 & 57.5 & - & - \\
VideoTree \cite{videotree} & GPT-4 & avg. 62 & - & - & 54.2 & - & 61.1 & - \\
Q-Frame (GPT-4o) \cite{zhang2025q} & GPT-4o & 8 & 63.8 & \underline{69.9} & \underline{63.8} & 57.6 & - & 58.6 \\
VSLS (GPT-4o) \cite{guo2025logic} & GPT-4o & 32 & 71.9 & 61.9 & 55.2 & 63.0 & - & 63.4 \\
VSLS (InternVL2.5-78B) \cite{guo2025logic} & 78B & 8 & 59.0 & 57.5 & 57.7 & 58.1 & - & \textit{64.5} \\
\midrule
\multicolumn{9}{c}{\textbf{MaxInfo (ours)}} \\ 
\midrule
\textbf{MaxInfo} (InternVL2) & 1B & * & - & - & - & - & - & 43.9 \\ 
\textbf{MaxInfo} (Qwen2-VL) & 7B & * & 72.5 & 62.0 & 51.8 & 62.1 & 64.3 & 55.7 \\
\textbf{MaxInfo} (LLaVA-Video) & 7B & * & 74.6 & 63.3 & 54.6 & 64.2 & 63.7 & 61.5 \\
\textbf{MaxInfo} (LLaVA-Video) & 72B & * & \underline{80.2} & \textit{67.7} & \textit{62.7} & \underline{70.2} & \textbf{69.4} & \underline{64.9} \\
\bottomrule
\end{tabular}
}
\end{table*}

\subsection{Fast vs.\ Slow Version Comparisons} \label{subsec:fast-slow}

As shown in Table~\ref{slow-fast-version}, the slow version outperforms the fast version in most cases, especially when processing long videos. Its initial oversampling mechanism provides MaxInfo with a richer selection space, which significantly improves performance. However, experiments have also shown that the fast version may outperform the slow version in certain benchmarks or when the number of initial samples is small. The MaxInfo block time is very short, so latency is almost negligible and more details on latency and GPUs memory and time.


\begin{table*}[htb!]
\caption{Performance of Fast vs. Slow MaxInfo variants on LLaVA-Video-7B (LongVideoBench). I/O: input/output frames. All results are reproduced, except those marked with *, which are copied from the corresponding papers \cite{llava-video}.}
\label{slow-fast-version}
\centering
\small
\setlength{\tabcolsep}{6pt}
\resizebox{0.95\textwidth}{!}{
\begin{tabular}{lccccc}
\toprule
Model & I max. frames & O frame range & Avg. frames & Encoding + MaxInfo Time & Acc. \\
\midrule
InternVL2 1B \cite{internvl} & 32 & - & 32 & - & 43.38 \\
\textbf{+ MaxInfo}$_{fast}$ & 32 & (1, 32) & 30 & 0.179 + 0.0126 s & \underline{43.60 }{\scriptsize \color{positive}+0.22\%} \\
\textbf{+ MaxInfo}$_{slow}$ & 64 & (1, 16) & 16 & 0.339 + 0.0215 s & \textbf{44.65} {\scriptsize \color{positive}+1.05\%} \\
\midrule
InternVL2 2B \cite{internvl} & 32 & - & 32 & - & \underline{46.97} \\
\textbf{+ MaxInfo}$_{fast}$ & 32 & (1, 32) & 30 & 0.179 + 0.0126 s & \textbf{47.27} {\scriptsize \color{positive}+0.30\%} \\
\textbf{+ MaxInfo}$_{slow}$ & 64 & (1, 16) & 16 & 0.339 + 0.0215 s & 46.82 {\scriptsize \color{red}-0.15\%} \\
\midrule
LLaVA-Video 7B \cite{llava-video} & 64 & - & 64 & - & 58.2$^*$ \\
\textbf{+ MaxInfo}$_{fast}$ & 64 & (1, 64) & 52 & 0.339 + 0.0215 s & \underline{60.21} {\scriptsize \color{positive}+2.01\%} \\
\textbf{+ MaxInfo}$_{slow}$ & 128 & (1, 64) & 58 & 0.624 + 0.0421 s & \textbf{61.48} {\scriptsize \color{positive}+3.28\%} \\
\midrule
LLaVA-Video 72B \cite{llava-video} & 64 & - & 64 & - & 61.9$^*$ \\
\textbf{+ MaxInfo}$_{fast}$ & 64 & (1, 64) & 52 & 0.339 + 0.0215 s & \textbf{65.37} {\scriptsize \color{positive}+3.47\%} \\
\textbf{+ MaxInfo}$_{slow}$ & 128 & (1, 64) & 58 & 0.624 + 0.0421 s & \underline{64.92} {\scriptsize \color{positive}+3.02\%} \\
\bottomrule
\end{tabular}
}
\end{table*}

\subsection{Ablation Study}

To further evaluate the effectiveness of MaxInfo, we conducted a series of ablation experiments focusing on two key aspects: the impact of the choice of visual encoder and the influence of key hyperparameters in the MaxInfo module, particularly tolerance and rank.

\subsubsection{Vision Encoder Impact}

As shown in Table~\ref{different-encoder}, we evaluated CLIP, DINOv2, and SigLIP as visual encoders. DINOv2 performed comparably to CLIP, despite lacking vision-language alignment. However, SigLIP outperformed both CLIP and DINOv2, likely due to its stronger language-vision connectivity. Interestingly, the larger SigLIP model underperformed, suggesting that more complex encoders require careful hyperparameter tuning (e.g., tolerance and rank) for optimal frame selection.

\begin{table}[htb!]
\caption{Visual encoder ablation for MaxInfo. Best in \textbf{bold}, second-best \underline{underlined}. Based on LLaVA-Video-7B \cite{llava-video}.}
\label{different-encoder}
\centering
\small
\begin{tabular}{lccc}
\toprule
\textbf{Model}  & \textbf{Visual Encoder} & \textbf{Param.} & \textbf{Acc.} \\
\midrule
LLaVA-Video \cite{llava-video} & CLIP-ViT-Large & 427.9M & 58.94 \\
LLaVA-Video \cite{llava-video} & CLIP-ViT-Base & 149.6M & 58.79 \\
LLaVA-Video \cite{llava-video} & DINO2-base & 86.6M & 58.94 \\
LLaVA-Video \cite{llava-video} & DINO2-large & 304.4M & 58.86 \\
LLaVA-Video \cite{llava-video} & SigLIP-base-224 & 203.2M & \textbf{59.76} \\
LLaVA-Video \cite{llava-video} & SigLIP-base-384 & 878.0M & \underline{59.24} \\
\bottomrule
\end{tabular}
\end{table}

\subsubsection{Impact of MaxInfo Block Hyperparameters} \label{4-4-2-hyper}

This section examines the effect of MaxInfo parameters (rank R and tolerance Tol) as well as the initial number of samples on the final model accuracy.

\begin{figure}[htb!]
  \centering
  \includegraphics[width=0.8\linewidth]{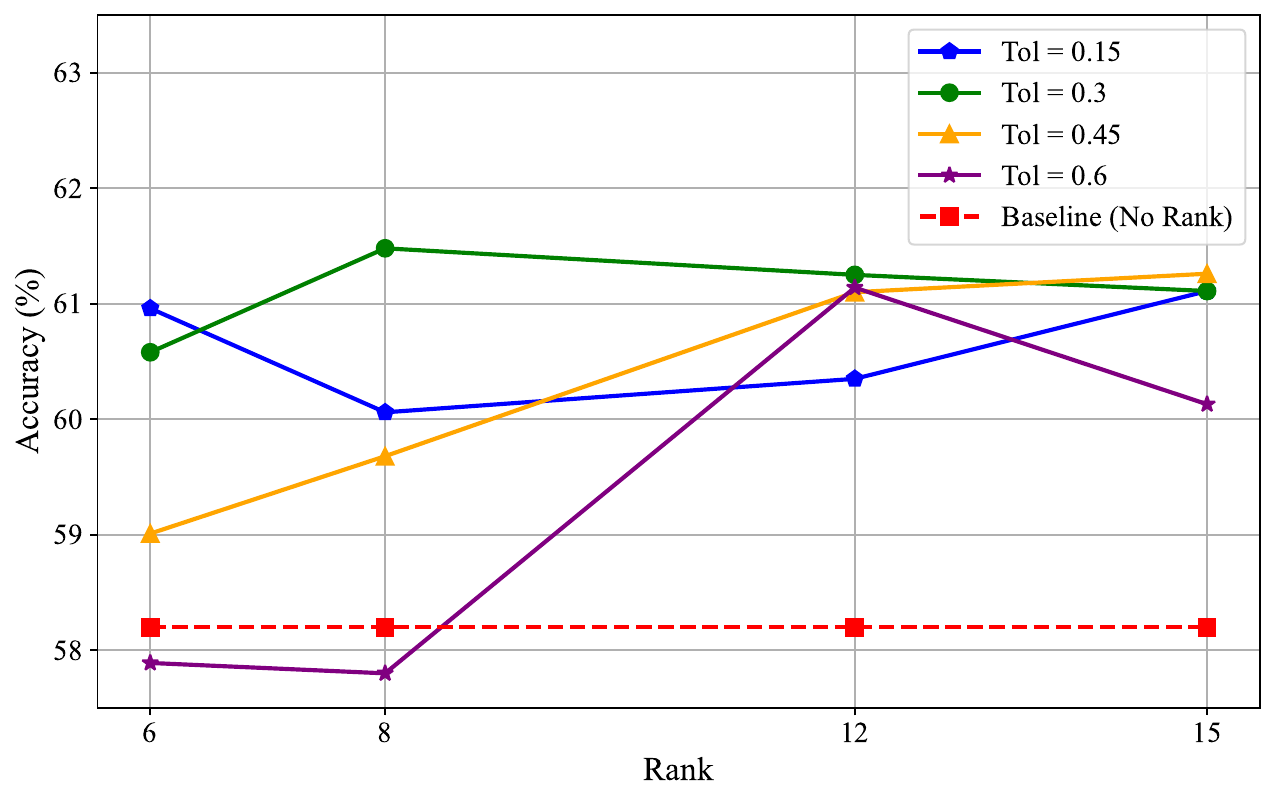}
  \caption{Effect of initial sampling on MaxInfo performance for LlaVa-Video 7B model.}
  \label{un-fixed-r-tol}
\end{figure}

\textbf{Effect of Different Tolerances and Ranks on the Model with Fixed Sampling.} We conducted a series of tests with fixed benchmarks, model settings, and an initial pool of frames (e.g., \(n^*=96\)). As shown in Figure~\ref{un-fixed-r-tol}, the best observed result achieved a \textbf{3.3\% improvement} over the base LLaVA-Video-7B model with \(\text{Tol} = 0.3\) and \(\text{R} = 8\). From these experiments, we derived the following guidelines:

\begin{enumerate}[leftmargin=*]
    \vspace{-0.03in}
    \item Performance Sensitivity: Our experiments show that performance is most sensitive to \(\text{Tol}\) values in the range \(\text{Tol} \in [0.15, 0.60]\). Beyond this range, improvements plateau or regress due to over-pruning or under-pruning.
    \vspace{-0.03in}
    \item Convergence: The model will converge when setting $\text{R} \in [12, 15], \text{Tol} \in [0.3, 0.45]$, this means that our choice of hyperparameters is not intractable.
\end{enumerate}

\begin{figure}[htb!]
  \centering
  \includegraphics[width=0.8\linewidth]{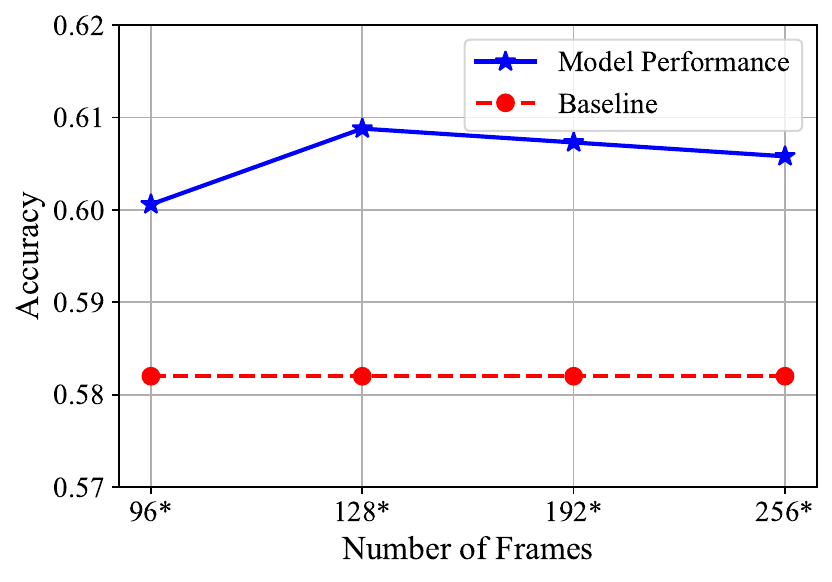}
  \caption{Effect of Initial Sampling on MaxInfo. Starting from $n^*$ sampled frames, the MaxInfo Block selects up to 64 informative frames for further processing.}
  \label{fixed-r-tol}
\end{figure}

\textbf{Effect of Sampling on Accuracy} We analyzed how varying the initial number of sampled frames impacts accuracy, while keeping the hyperparameters fixed (R=8, Tol=0.15). As shown in Figure~\ref{fixed-r-tol}, accuracy initially improves with the addition of more frames, but beyond a certain threshold, it begins to plateau or slightly decline. Our key observations are as follows:

\begin{enumerate}[leftmargin=*]
    \vspace{-0.03in}
    \item Increasing the initial number of frames provides more diverse information, enabling MaxVol to select better keyframes, but the information converges.
    \vspace{-0.03in}
    \item There exists an optimal trade-off between the initial frame count and computational cost. In our tests, 128 as initial frames yielded the best accuracy for the LLaVA-Video-7B model.
\end{enumerate}

\subsubsection{Case Study} \label{case-study}

\begin{figure*}[htb!]
    \centering
    \includegraphics[width=1\linewidth]{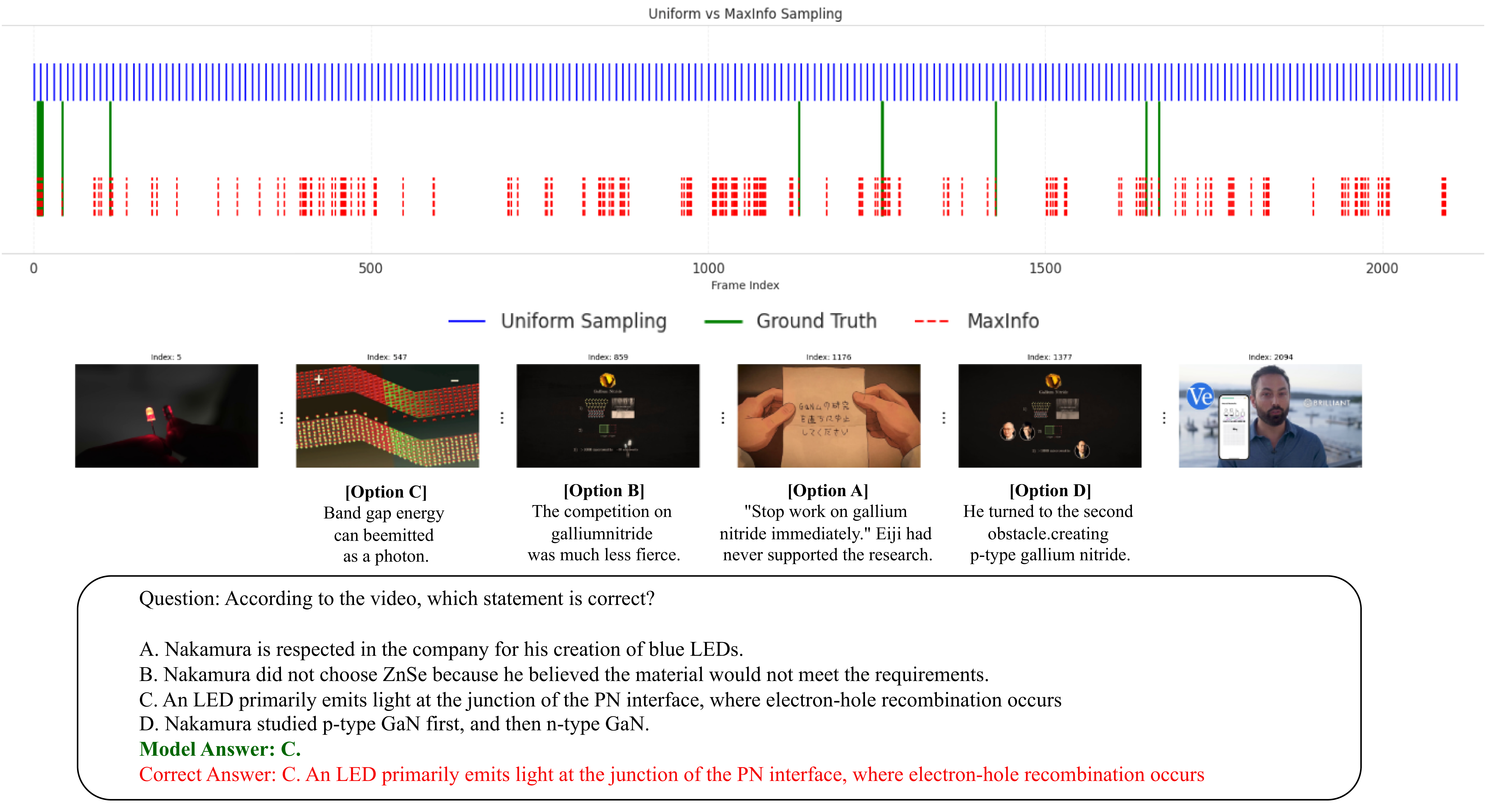}
    \caption{Qualitative: (a) MaxInfo vs Uniform Sampling with GT-aligned frames; (b) CLIP scores show MaxInfo's answer coverage in single samples.}
    \label{qualitative}
\end{figure*}

Figure~\ref{qualitative} presents a long-video example from the Video-MME \cite{video-mme} dataset, qualitatively illustrating the effectiveness of the proposed MaxInfo Block. We compare frame selection by MaxInfo Block with Uniform Sampling and observe that the keyframes chosen by MaxInfo are more closely aligned with the manually annotated Ground Truth–related frames. 


\section{Conclusion}

In this work, we introduced \textbf{MaxInfo} -- a training-free method for selecting the most informative frames from videos, improving VLLMs inference. Our results consistently demonstrate that informative frame selection outperforms uniform sampling, leading to improved performance of state of the art VLLMs (LLaVA-Video, InternVL and Qwen2-VL with different sizes) across multiple benchmarks. For example, MaxInfo achieves a 3.28\% improvement on LongVideoBench and a 6.4\% improvement on EgoSchema for LLaVA-Video-7B. The Slow Version of MaxInfo improves LLaVA-Video-72B performance by 3.47\% on LongVideoBench.

Beyond demonstrating empirical gains, we believe our work will encourage the community to focus more on frame selection strategies, an often-overlooked aspect of video understanding. Additionally, we have shown that even minimal refinements, such as chunk-wise MaxVol in our Scene-Aware MaxInfo, can further enhance results, demonstrating that simple adjustments can lead to meaningful improvements.

Finally, we hypothesize that training VLLMs with informative frame sampling, rather than simple uniform frame selection, could further enhance their capabilities when later used with inference-time MaxInfo techniques. We hope this work serves as a foundation for future research into more efficient, information-aware video sampling strategies for large-scale multimodal learning.

\section{Acknowledgments}
Innopolis University authors were supported by the Research Center of the Artificial Intelligence Institute at Innopolis University. Financial support was provided by the Ministry of Economic Development of the Russian Federation (No. 25-139-66879-1-0003).


{
    \small
    \bibliographystyle{ieeenat_fullname}
    \bibliography{main}

@article{Rectmax,
  title={Rectangular maximum-volume submatrices and their applications},
  author={Mikhalev, Aleksandr and Oseledets, Ivan V},
  journal={Linear Algebra and its Applications},
  volume={538},
  pages={187--211},
  year={2018},
  publisher={Elsevier}
}

@article{MOI,
  title={Feature selection in MLPs and SVMs based on maximum output information},
  author={Sindhwani, Vikas and Rakshit, Subrata and Deodhare, Dipti and Erdogmus, Deniz and Principe, Jos{\'e} Carlos and Niyogi, Partha},
  journal={IEEE transactions on neural networks},
  volume={15},
  number={4},
  pages={937--948},
  year={2004},
  publisher={IEEE}
}

@article{MMD,
  title={Feature selection by maximum marginal diversity},
  author={Vasconcelos, Nuno},
  journal={Advances in neural information processing systems},
  volume={15},
  year={2002}
}

@article{peng2005feature,
  title={Feature selection based on mutual information criteria of max-dependency, max-relevance, and min-redundancy},
  author={Peng, Hanchuan and Long, Fuhui and Ding, Chris},
  journal={IEEE Transactions on pattern analysis and machine intelligence},
  volume={27},
  number={8},
  pages={1226--1238},
  year={2005},
  publisher={IEEE}
}

@incollection{goreinov2010find,
  title={How to find a good submatrix},
  author={Goreinov, Sergei A and Oseledets, Ivan V and Savostyanov, Dimitry V and Tyrtyshnikov, Eugene E and Zamarashkin, Nikolay L},
  booktitle={Matrix Methods: Theory, Algorithms And Applications: Dedicated to the Memory of Gene Golub},
  pages={247--256},
  year={2010},
  publisher={World Scientific}
}

@article{sozykin2022ttopt,
  title={TTOpt: A maximum volume quantized tensor train-based optimization and its application to reinforcement learning},
  author={Sozykin, Konstantin and Chertkov, Andrei and Schutski, Roman and Phan, Anh-Huy and Cichocki, Andrzej S and Oseledets, Ivan},
  journal={Advances in Neural Information Processing Systems},
  volume={35},
  pages={26052--26065},
  year={2022}
}

@article{video-lavit,
  title={Video-lavit: Unified video-language pre-training with decoupled visual-motional tokenization},
  author={Jin, Yang and Sun, Zhicheng and Xu, Kun and Chen, Liwei and Jiang, Hao and Huang, Quzhe and Song, Chengru and Liu, Yuliang and Zhang, Di and Song, Yang and others},
  journal={arXiv preprint arXiv:2402.03161},
  year={2024}
}

@article{qwen2-vl,
  title={Qwen2-vl: Enhancing vision-language model's perception of the world at any resolution},
  author={Wang, Peng and Bai, Shuai and Tan, Sinan and Wang, Shijie and Fan, Zhihao and Bai, Jinze and Chen, Keqin and Liu, Xuejing and Wang, Jialin and Ge, Wenbin and others},
  journal={arXiv preprint arXiv:2409.12191},
  year={2024}
}

@article{llava-video,
  title={Video instruction tuning with synthetic data},
  author={Zhang, Yuanhan and Wu, Jinming and Li, Wei and Li, Bo and Ma, Zejun and Liu, Ziwei and Li, Chunyuan},
  journal={arXiv preprint arXiv:2410.02713},
  year={2024}
}

@article{slowfast,
  title={Slowfast-llava: A strong training-free baseline for video large language models},
  author={Xu, Mingze and Gao, Mingfei and Gan, Zhe and Chen, Hong-You and Lai, Zhengfeng and Gang, Haiming and Kang, Kai and Dehghan, Afshin},
  journal={arXiv preprint arXiv:2407.15841},
  year={2024}
}

@article{llava,
  title={Visual instruction tuning},
  author={Liu, Haotian and Li, Chunyuan and Wu, Qingyang and Lee, Yong Jae},
  journal={Advances in neural information processing systems},
  volume={36},
  year={2024}
}

@inproceedings{song2024moviechat,
  title={Moviechat: From dense token to sparse memory for long video understanding},
  author={Song, Enxin and Chai, Wenhao and Wang, Guanhong and Zhang, Yucheng and Zhou, Haoyang and Wu, Feiyang and Chi, Haozhe and Guo, Xun and Ye, Tian and Zhang, Yanting and others},
  booktitle={Proceedings of the IEEE/CVF Conference on Computer Vision and Pattern Recognition},
  pages={18221--18232},
  year={2024}
}

@article{video-mme,
  title={Video-mme: The first-ever comprehensive evaluation benchmark of multi-modal llms in video analysis},
  author={Fu, Chaoyou and Dai, Yuhan and Luo, Yongdong and Li, Lei and Ren, Shuhuai and Zhang, Renrui and Wang, Zihan and Zhou, Chenyu and Shen, Yunhang and Zhang, Mengdan and others},
  journal={arXiv preprint arXiv:2405.21075},
  year={2024}
}

@inproceedings{clip,
  title={Learning transferable visual models from natural language supervision},
  author={Radford, Alec and Kim, Jong Wook and Hallacy, Chris and Ramesh, Aditya and Goh, Gabriel and Agarwal, Sandhini and Sastry, Girish and Askell, Amanda and Mishkin, Pamela and Clark, Jack and others},
  booktitle={International conference on machine learning},
  pages={8748--8763},
  year={2021},
  organization={PMLR}
}

@inproceedings{blip2,
  title={Blip-2: Bootstrapping language-image pre-training with frozen image encoders and large language models},
  author={Li, Junnan and Li, Dongxu and Savarese, Silvio and Hoi, Steven},
  booktitle={International conference on machine learning},
  pages={19730--19742},
  year={2023},
  organization={PMLR}
}

@article{longvu,
  title={Longvu: Spatiotemporal adaptive compression for long video-language understanding},
  author={Shen, Xiaoqian and Xiong, Yunyang and Zhao, Changsheng and Wu, Lemeng and Chen, Jun and Zhu, Chenchen and Liu, Zechun and Xiao, Fanyi and Varadarajan, Balakrishnan and Bordes, Florian and others},
  journal={arXiv preprint arXiv:2410.17434},
  year={2024}
}

@article{longvideobench,
  title={Longvideobench: A benchmark for long-context interleaved video-language understanding},
  author={Wu, Haoning and Li, Dongxu and Chen, Bei and Li, Junnan},
  journal={arXiv preprint arXiv:2407.15754},
  year={2024}
}

@article{mlvu,
  title={MLVU: A Comprehensive Benchmark for Multi-Task Long Video Understanding},
  author={Zhou, Junjie and Shu, Yan and Zhao, Bo and Wu, Boya and Xiao, Shitao and Yang, Xi and Xiong, Yongping and Zhang, Bo and Huang, Tiejun and Liu, Zheng},
  journal={arXiv preprint arXiv:2406.04264},
  year={2024}
}

@article{team2024gemini,
  title={Gemini 1.5: Unlocking multimodal understanding across millions of tokens of context},
  author={Team, Gemini and Georgiev, Petko and Lei, Ving Ian and Burnell, Ryan and Bai, Libin and Gulati, Anmol and Tanzer, Garrett and Vincent, Damien and Pan, Zhufeng and Wang, Shibo and others},
  journal={arXiv preprint arXiv:2403.05530},
  year={2024}
}

@article{llama3,
  title={The llama 3 herd of models},
  author={Dubey, Abhimanyu and Jauhri, Abhinav and Pandey, Abhinav and Kadian, Abhishek and Al-Dahle, Ahmad and Letman, Aiesha and Mathur, Akhil and Schelten, Alan and Yang, Amy and Fan, Angela and others},
  journal={arXiv preprint arXiv:2407.21783},
  year={2024}
}

@article{qwen2,
  title={Qwen2. 5 Technical Report},
  author={Yang, An and Yang, Baosong and Zhang, Beichen and Hui, Binyuan and Zheng, Bo and Yu, Bowen and Li, Chengyuan and Liu, Dayiheng and Huang, Fei and Wei, Haoran and others},
  journal={arXiv preprint arXiv:2412.15115},
  year={2024}
}

@article{gpt4,
  title={Gpt-4 technical report},
  author={Achiam, Josh and Adler, Steven and Agarwal, Sandhini and Ahmad, Lama and Akkaya, Ilge and Aleman, Florencia Leoni and Almeida, Diogo and Altenschmidt, Janko and Altman, Sam and Anadkat, Shyamal and others},
  journal={arXiv preprint arXiv:2303.08774},
  year={2023}
}

@article{omnifusion,
  title={Omnifusion technical report},
  author={Goncharova, Elizaveta and Razzhigaev, Anton and Mikhalchuk, Matvey and Kurkin, Maxim and Abdullaeva, Irina and Skripkin, Matvey and Oseledets, Ivan and Dimitrov, Denis and Kuznetsov, Andrey},
  journal={arXiv preprint arXiv:2404.06212},
  year={2024}
}

@inproceedings{internvl,
  title={Internvl: Scaling up vision foundation models and aligning for generic visual-linguistic tasks},
  author={Chen, Zhe and Wu, Jiannan and Wang, Wenhai and Su, Weijie and Chen, Guo and Xing, Sen and Zhong, Muyan and Zhang, Qinglong and Zhu, Xizhou and Lu, Lewei and others},
  booktitle={Proceedings of the IEEE/CVF Conference on Computer Vision and Pattern Recognition},
  pages={24185--24198},
  year={2024}
}

@inproceedings{Chat-univi,
  title={Chat-univi: Unified visual representation empowers large language models with image and video understanding},
  author={Jin, Peng and Takanobu, Ryuichi and Zhang, Wancai and Cao, Xiaochun and Yuan, Li},
  booktitle={Proceedings of the IEEE/CVF Conference on Computer Vision and Pattern Recognition},
  pages={13700--13710},
  year={2024}
}

@inproceedings{tang2025adaptive,
  title={Adaptive keyframe sampling for long video understanding},
  author={Tang, Xi and Qiu, Jihao and Xie, Lingxi and Tian, Yunjie and Jiao, Jianbin and Ye, Qixiang},
  booktitle={Proceedings of the Computer Vision and Pattern Recognition Conference},
  pages={29118--29128},
  year={2025}
}

@article{chen2024sharegpt4video,
  title={Sharegpt4video: Improving video understanding and generation with better captions},
  author={Chen, Lin and Wei, Xilin and Li, Jinsong and Dong, Xiaoyi and Zhang, Pan and Zang, Yuhang and Chen, Zehui and Duan, Haodong and Tang, Zhenyu and Yuan, Li and others},
  journal={Advances in Neural Information Processing Systems},
  volume={37},
  pages={19472--19495},
  year={2024}
}

@article{cheng2024videollama,
  title={Videollama 2: Advancing spatial-temporal modeling and audio understanding in video-llms},
  author={Cheng, Zesen and Leng, Sicong and Zhang, Hang and Xin, Yifei and Li, Xin and Chen, Guanzheng and Zhu, Yongxin and Zhang, Wenqi and Luo, Ziyang and Zhao, Deli and others},
  journal={arXiv preprint arXiv:2406.07476},
  year={2024}
}

@inproceedings{liu2025bolt,
  title={BOLT: Boost Large Vision-Language Model Without Training for Long-form Video Understanding},
  author={Liu, Shuming and Zhao, Chen and Xu, Tianqi and Ghanem, Bernard},
  booktitle={Proceedings of the Computer Vision and Pattern Recognition Conference},
  pages={3318--3327},
  year={2025}
}

@article{wang2025internvl3,
  title={Internvl3. 5: Advancing open-source multimodal models in versatility, reasoning, and efficiency},
  author={Wang, Weiyun and Gao, Zhangwei and Gu, Lixin and Pu, Hengjun and Cui, Long and Wei, Xingguang and Liu, Zhaoyang and Jing, Linglin and Ye, Shenglong and Shao, Jie and others},
  journal={arXiv preprint arXiv:2508.18265},
  year={2025}
}

@inproceedings{hu2025m,
  title={M-LLM based video frame selection for efficient video understanding},
  author={Hu, Kai and Gao, Feng and Nie, Xiaohan and Zhou, Peng and Tran, Son and Neiman, Tal and Wang, Lingyun and Shah, Mubarak and Hamid, Raffay and Yin, Bing and others},
  booktitle={Proceedings of the Computer Vision and Pattern Recognition Conference},
  pages={13702--13712},
  year={2025}
}

@article{videotree,
  title={VideoTree: Adaptive Tree-based Video Representation for LLM Reasoning on Long Videos},
  author={Wang, Ziyang and Yu, Shoubin and Stengel-Eskin, Elias and Yoon, Jaehong and Cheng, Feng and Bertasius, Gedas and Bansal, Mohit},
  journal={arXiv preprint arXiv:2405.19209},
  year={2024}
}

@inproceedings{dino,
  title={Emerging properties in self-supervised vision transformers},
  author={Caron, Mathilde and Touvron, Hugo and Misra, Ishan and J{\'e}gou, Herv{\'e} and Mairal, Julien and Bojanowski, Piotr and Joulin, Armand},
  booktitle={Proceedings of the IEEE/CVF international conference on computer vision},
  pages={9650--9660},
  year={2021}
}

@inproceedings{Siglip,
  title={Sigmoid loss for language image pre-training},
  author={Zhai, Xiaohua and Mustafa, Basil and Kolesnikov, Alexander and Beyer, Lucas},
  booktitle={Proceedings of the IEEE/CVF International Conference on Computer Vision},
  pages={11975--11986},
  year={2023}
}

@inproceedings{mvbench,
  title={Mvbench: A comprehensive multi-modal video understanding benchmark},
  author={Li, Kunchang and Wang, Yali and He, Yinan and Li, Yizhuo and Wang, Yi and Liu, Yi and Wang, Zun and Xu, Jilan and Chen, Guo and Luo, Ping and others},
  booktitle={Proceedings of the IEEE/CVF Conference on Computer Vision and Pattern Recognition},
  pages={22195--22206},
  year={2024}
}

@article{xenos2024vllms,
  title={VLLMs Provide Better Context for Emotion Understanding Through Common Sense Reasoning},
  author={Xenos, Alexandros and Foteinopoulou, Niki Maria and Ntinou, Ioanna and Patras, Ioannis and Tzimiropoulos, Georgios},
  journal={arXiv preprint arXiv:2404.07078},
  year={2024}
}

@article{jiang2023mistral,
  title={Mistral 7B},
  author={Jiang, Albert Q and Sablayrolles, Alexandre and Mensch, Arthur and Bamford, Chris and Chaplot, Devendra Singh and Casas, Diego de las and Bressand, Florian and Lengyel, Gianna and Lample, Guillaume and Saulnier, Lucile and others},
  journal={arXiv preprint arXiv:2310.06825},
  year={2023}
}

@article{gpt-3,
  title={GPT-3: Its nature, scope, limits, and consequences},
  author={Floridi, Luciano and Chiriatti, Massimo},
  journal={Minds and Machines},
  volume={30},
  pages={681--694},
  year={2020},
  publisher={Springer}
}

@article{llama2,
  title={Llama 2: Open foundation and fine-tuned chat models},
  author={Touvron, Hugo and Martin, Louis and Stone, Kevin and Albert, Peter and Almahairi, Amjad and Babaei, Yasmine and Bashlykov, Nikolay and Batra, Soumya and Bhargava, Prajjwal and Bhosale, Shruti and others},
  journal={arXiv preprint arXiv:2307.09288},
  year={2023}
}

@article{qwen,
  title={Qwen technical report},
  author={Bai, Jinze and Bai, Shuai and Chu, Yunfei and Cui, Zeyu and Dang, Kai and Deng, Xiaodong and Fan, Yang and Ge, Wenbin and Han, Yu and Huang, Fei and others},
  journal={arXiv preprint arXiv:2309.16609},
  year={2023}
}

@misc{zhang2024lmmseval,
    title={LMMs-Eval: Reality Check on the Evaluation of Large Multimodal Models},
    author={Kaichen Zhang and Bo Li and Peiyuan Zhang and Fanyi Pu and Joshua Adrian Cahyono and Kairui Hu and Shuai Liu and Yuanhan Zhang and Jingkang Yang and Chunyuan Li and Ziwei Liu},
    year={2024},
    eprint={2407.12772},
    archivePrefix={arXiv},
    primaryClass={cs.CL}
}

@article{egoschema,
  title={Egoschema: A diagnostic benchmark for very long-form video language understanding},
  author={Mangalam, Karttikeya and Akshulakov, Raiymbek and Malik, Jitendra},
  journal={Advances in Neural Information Processing Systems},
  volume={36},
  pages={46212--46244},
  year={2023}
}

@article{cheng2024videollama2,
  title={VideoLLaMA 2: Advancing Spatial-Temporal Modeling and Audio Understanding in Video-LLMs},
  author={Cheng, Zesen and Leng, Sicong and Zhang, Hang and Xin, Yifei and Li, Xin and Chen, Guanzheng and Zhu, Yongxin and Zhang, Wenqi and Luo, Ziyang and Zhao, Deli and Bing, Lidong},
  journal={arXiv preprint arXiv:2406.07476},
  year={2024},
  url={https://arxiv.org/abs/2406.07476}
}

@article{yao2024minicpmv,
  title={MiniCPM-V: A GPT-4V Level MLLM on Your Phone},
  author={Yao, Yuan and Yu, Tianyu and Zhang, Ao and Wang, Chongyi and Cui, Junbo and Zhu, Hongji and Cai, Tianchi and Li, Haoyu and Zhao, Weilin and He, Zhihui and others},
  journal={arXiv preprint arXiv:2408.01800},
  year={2024}
}

@article{xue2024longvila,
  title={Longvila: Scaling long-context visual language models for long videos},
  author={Xue, Fuzhao and Chen, Yukang and Li, Dacheng and Hu, Qinghao and Zhu, Ligeng and Li, Xiuyu and Fang, Yunhao and Tang, Haotian and Yang, Shang and Liu, Zhijian and others},
  journal={arXiv preprint arXiv:2408.10188},
  year={2024}
}

@article{guo2025logic,
  title={Logic-in-frames: Dynamic keyframe search via visual semantic-logical verification for long video understanding},
  author={Guo, Weiyu and Chen, Ziyang and Wang, Shaoguang and He, Jianxiang and Xu, Yijie and Ye, Jinhui and Sun, Ying and Xiong, Hui},
  journal={arXiv preprint arXiv:2503.13139},
  year={2025}
}

@article{wang2025adaretake,
  title={Adaretake: Adaptive redundancy reduction to perceive longer for video-language understanding},
  author={Wang, Xiao and Si, Qingyi and Wu, Jianlong and Zhu, Shiyu and Cao, Li and Nie, Liqiang},
  journal={arXiv preprint arXiv:2503.12559},
  year={2025}
}

@article{zhang2025q,
  title={Q-Frame: Query-aware Frame Selection and Multi-Resolution Adaptation for Video-LLMs},
  author={Zhang, Shaojie and Yang, Jiahui and Yin, Jianqin and Luo, Zhenbo and Luan, Jian},
  journal={arXiv preprint arXiv:2506.22139},
  year={2025}
}

@article{yao2025generative,
  title={Generative Frame Sampler for Long Video Understanding},
  author={Yao, Linli and Wu, Haoning and Ouyang, Kun and Zhang, Yuanxing and Xiong, Caiming and Chen, Bei and Sun, Xu and Li, Junnan},
  journal={arXiv preprint arXiv:2503.09146},
  year={2025}
}

@article{cheng2025scaling,
  title={Scaling Video-Language Models to 10K Frames via Hierarchical Differential Distillation},
  author={Cheng, Chuanqi and Guan, Jian and Wu, Wei and Yan, Rui},
  journal={arXiv preprint arXiv:2504.02438},
  year={2025}
}

@article{zhang2024long,
  title={Long context transfer from language to vision},
  author={Zhang, Peiyuan and Zhang, Kaichen and Li, Bo and Zeng, Guangtao and Yang, Jingkang and Zhang, Yuanhan and Wang, Ziyue and Tan, Haoran and Li, Chunyuan and Liu, Ziwei},
  journal={arXiv preprint arXiv:2406.16852},
  year={2024}
}

@article{yu2024frame,
  title={Frame-voyager: Learning to query frames for video large language models},
  author={Yu, Sicheng and Jin, Chengkai and Wang, Huanyu and Chen, Zhenghao and Jin, Sheng and Zuo, Zhongrong and Xu, Xiaolei and Sun, Zhenbang and Zhang, Bingni and Wu, Jiawei and others},
  journal={arXiv preprint arXiv:2410.03226},
  year={2024}
}

@article{openai2024gpt4o,
  title={GPT-4o System Card},
  author={OpenAI},
  journal={arXiv preprint arXiv:2410.21276},
  year={2024},
  url={https://arxiv.org/abs/2410.21276}
}
}

\newpage
\appendix

\section{Implementation Details} \label{details}

 We mostly focus on longer videos because better frame selection plays a bigger role in longer, more complex videos, whereas shorter ones intuitively work well with uniform sampling due to their lower information content and complexity.

We ensured that the resulting sequence length of a set of visual and textual tokens did not exceed the maximum sequence length for this LLM. When evaluating models using MaxInfo, we limited the number of selected frames so that they did not exceed the maximum allowed for the context of the estimated VLLM. For the evaluation on all benchmarks, we have set the generation temperature to 0.

For the general multiple-choice question-answering evaluation, we follow the official guidelines to construct the instructions using the provided questions and options. We added a prompt to the question and options like \textit{"Respond with only the letter (A, B, C, or D) of the correct option."} for LongVideoBench \cite{longvideobench}, Video-MME \cite{video-mme}, MLVU \cite{mlvu} and MVBench \cite{mvbench} or \textit{"Answer with the option's letter from the given choices directly and only give the best option."} for EgoSchema \cite{egoschema}. We follow the original benchmarks setup to calculate the final scores, and we also align our evaluation protocols with other evaluation toolkits, such as lmms-eval \cite{zhang2024lmmseval}.

To ensure the reproducibility of our results, we have included the main hyperparameters used for all benchmarks and estimated models in the results tables, such as tolerance and rank for the MaxInfo algorithm, the number of sampled frames, and the number of initial frames (before MaxInfo).

\section{Additional Experiments and Details}\label{sec:appendix-details}

To further assess the impact of MaxInfo, we evaluate its performance with an additional set of models \cite{wang2025internvl3}, \cite{yao2024minicpmv} on the LongVideoBench and Video-MME benchmarks. 

\subsection{Applying MaxInfo to recent models}

The results in Table~\ref{new-models} show that MaxInfo consistently improves model performance across both benchmarks, suggesting that precise frame selection is particularly important for long-video tasks.

\begin{table}[htb!]
\caption{Adaptation of MaxInfo to current new long video understanding models}
\label{new-models}
\centering
\small
\resizebox{\linewidth}{!}{
\begin{tabular}{lcccc}
\toprule
\textbf{Model}  & \textbf{Size} & \textbf{Frame Interval} & \textbf{Avg. frames} & \textbf{LongVideoBench.} \\
\midrule
MiniCPM \cite{yao2024minicpmv} & 9B & 128 & 128 & 56.17 \\
\textbf{+ MaxInfo} & 9B & [8, 82] & 56 & 59.61 \\
\rowcolor{lightlightgray} $\triangle$ & & & & \color{positive}+3.44 \\

InternVL3.5 \cite{wang2025internvl3} & 1B & 16 & 16 & 47.7 \\
\textbf{+ MaxInfo} & 1B & [1, 16] & 16 & 49.0 \\
\rowcolor{lightlightgray} $\triangle$ & & & & \color{positive}+1.3 \\

InternVL3.5 \cite{wang2025internvl3} & 8B & 16 & 16 & 57.4 \\
\textbf{+ MaxInfo} & 8B & [1, 16] & 16 & 59.0 \\
\rowcolor{lightlightgray} $\triangle$ & & & & \color{positive}+1.6 \\

InternVL3.5 \cite{wang2025internvl3} & 38B & 16 & 16 & 60 \\
\textbf{+ MaxInfo} & 38B & [1, 16] & 16 & 61.6 \\
\rowcolor{lightlightgray} $\triangle$ & & & & \color{positive}+1.6 \\

\bottomrule
\end{tabular}
}
\end{table}

\subsection{Performance Analysis: MaxInfo vs. Uniform Sampling} \label{4-2-performance}

\begin{figure*}[ht]
  \centering
  \includegraphics[width=1\textwidth]{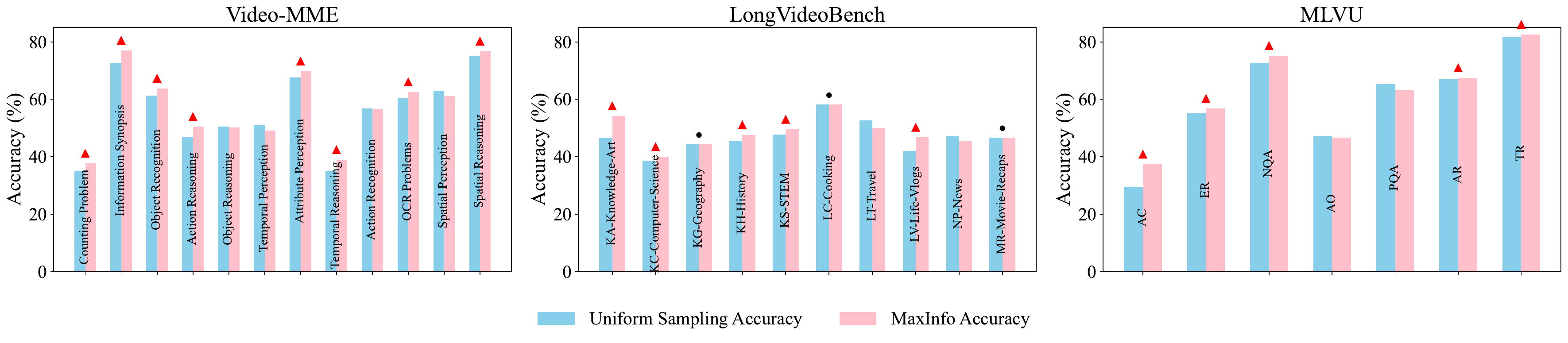}
  \caption{Accuracy comparison between Uniform Sampling and MaxInfo across three benchmarks.}
  \label{fig:combined-comparison}
\end{figure*}

To better understand the strengths and trade-offs of MaxInfo, we analyzed per-task accuracy across multiple benchmarks. Our results, as shown in Figure~\ref{fig:combined-comparison}, indicate that MaxInfo performs superiorly in high information density tasks such as counting, summarizing and spatial reasoning, while uniform sampling has a slight advantage in tasks that rely on temporal continuity, reflecting the key trade-off between information maximization and temporal consistency.

\subsection{Comparison with CLIP baseline} \label{clip-comparison}
As shown in Table~\ref{tab:longvideobench_maxinfo}, we compare the experimental results of two keyframe extraction strategies based on the QwenVL2-2B model on the LongVideoBench benchmark: the \textbf{CLIP-Based} thresholding method and the \textbf{MaxInfo} module method. Both methods extract the same number of frames in the initial phase, so the encoding time is kept the same, where the similarity threshold of the CLIP-Based method is set to 0.5. The results show that the MaxInfo module outperforms the CLIP-Based method in terms of the overall performance in keyframe selection. 

\begin{table}[htb!]
  \centering
  \caption{Performance comparison on LVBench.}
  \label{tab:longvideobench_maxinfo}
  \small
  \setlength{\tabcolsep}{8pt}
  \begin{tabular}{lcc}
    \toprule
    Model & Method & Accuracy \\
    \midrule
    QwenVL2-2B & CLIP-Based & 44.3 \\
    QwenVL2-2B & CLIP-Based + MaxInfo & 44.5 \\
    QwenVL2-2B & MaxInfo + CLIP-Based & 43.8 \\
    QwenVL2-2B & MaxInfo & \textbf{48.8} \\
    \bottomrule
  \end{tabular}
\end{table}

In addition, we also explored combining the CLIP-Based method with MaxInfo module. The experiments show that MaxInfo is able to improve the overall information quality of the input sequences, and its information maximization strategy plays a key role in frame selection, which further enhances the performance of the model. CLIP-Based loses a lot of semantic information, which can lead to performance degradation of the model.

In order to further evaluate whether MaxInfo will lose the key frames related to the problem, we compare MaxInfo with the Uniform Sampling method under the CLIP Score metric. The experimental results shown in Table~\ref{tab:clip_score_comparison} that MaxInfo does not miss the frames related to the semantics of the problem, and is able to retain the semantic relevance effectively.


\begin{table}[htb!]
  \centering
  \caption{CLIP score comparison between uniform and MaxInfo sampling.}
  \label{tab:clip_score_comparison}
  \small
  \setlength{\tabcolsep}{10pt}
  \begin{tabular}{lc}
    \toprule
    Sampling Method & CLIP-score \\
    \midrule
    Uniform & 0.37 \\
    MaxInfo & \textbf{0.39} \\
    \bottomrule
  \end{tabular}
\end{table}

\subsection{Qualitative comparison with uniform sampling} \label{key-maxinfo-vs}

We randomly selected 50 video samples in LongVideoBench and calculated the cosine similarity between the frames selected by MaxInfo and the cosine similarity between the frames obtained by uniform sampling.

\begin{figure}[htb!]
    \begin{center}
        \centerline{\includegraphics[width=0.4\textwidth]{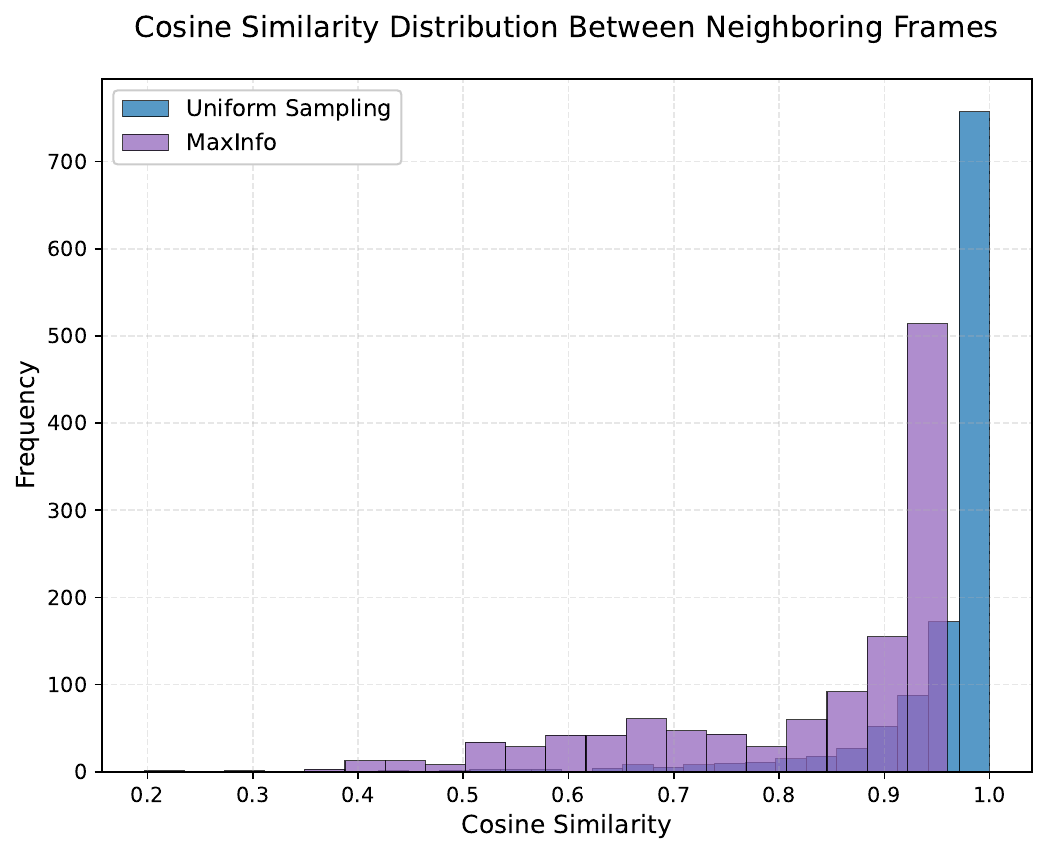}}
        \caption{Similarity distribution between neighbouring frames ($frame_i$ and $frame_{i+1}$).}
        \label{vs-distribution-maxinfo}
    \end{center}
\end{figure}

Figure~\ref{vs-distribution-maxinfo} shows the distribution of cosine similarity for the same number of frames. It is clear that MaxInfo produces a more diverse distribution like a low similarity offset compared to uniform sampling, highlighting its ability to capture more diverse visual content.

\begin{figure}[htb!]
    \begin{center}
        \centerline{\includegraphics[width=0.6\textwidth]{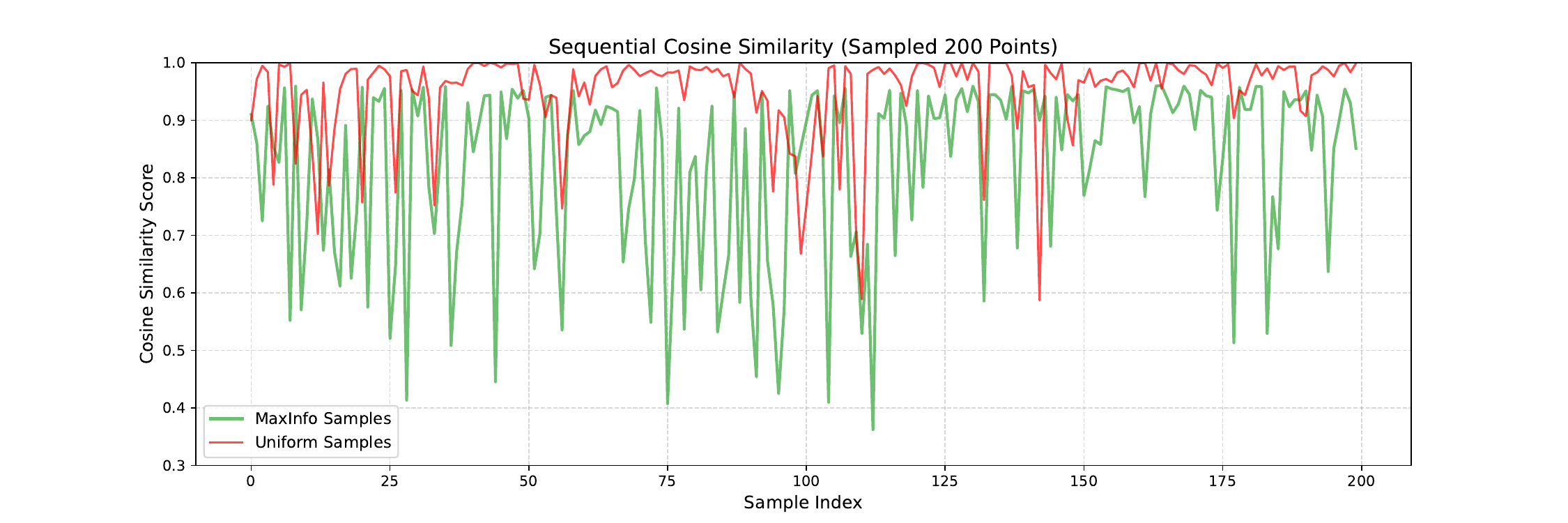}}
        \caption{CLIP similarity between neighboring frames selected by MaxInfo module.}
        \label{vs-cosine-maxinfo}
    \end{center}
\end{figure}

 As shown in Figure~\ref{vs-cosine-maxinfo}, we plotted 200 sampled data points to improve visual clarity. The results show that our MaxInfo module exhibits higher diversity in frame selection compared to uniform sampling.

\subsection{Computational Efficiency: Time and Memory Consumption} \label{time-and-mem}

When processing long videos, the LLM is the most resource-intensive component of VLLMs due to its parameter count and the quadratic complexity of attention with respect to input length. Since most of the context is occupied by visual tokens from frames, our MaxInfo method reduces this load by selecting keyframes. Importantly, MaxInfo requires minimal and constant memory and preprocessing time, independent of LLM size, and remains significantly lighter than uniformly sampling all frames.

\textbf{Time Complexity.} To evaluate the latency overhead of MaxInfo in practice, we measured its runtime with the Qwen2-VL model. As shown in Table~\ref{tab:runtime_components}, the runtime of MaxInfo is almost negligible compared to the inference time of the VLLM itself, confirming that MaxInfo is a lightweight and efficient frame selection mechanism. The initial VLLM time means the inference time of the 512 frames of information directly into the Qwen2-VL model. The frame count selected by MaxInfo Block is adaptive to the information content of the input. For near-static videos (low information density), MaxInfo drastically reduces the number of processed frames. Consequently, VLLMs + MaxInfo Block may achieve lower time compared to the initial VLLMs + MaxInfo Block configuration. The times reported in the table represent an upper bound; in practice, the reduced number of frames can lead to several-fold speedups on certain tasks.
 All experiments were conducted on an A100 GPU.

\begin{table}[htb!]
    \centering
    \caption{Runtime of different pipeline components, based on Qwen2-VL. Frame size = 512 (UP is Upper Bound).}
    \label{tab:runtime_components}
    \small
    \resizebox{0.95\linewidth}{!}{
    \begin{tabular}{lcccc}
        \toprule
        \textbf{Model Size} & \textbf{CLIP (s)} & \textbf{MaxVol (s)} & \textbf{VLLMs (s)} & \textbf{VLLMs + MaxInfo (UP)} \\
        \midrule
        2B & 0.296 & 0.0109 & 2.979 & $\leq$ 3.285 \\
        7B & 0.296 & 0.0109 & 5.372 & $\leq$ 5.679 \\
        72B & 0.296 & 0.0109 & 30.737 & $\leq$ 31.044 \\
        \bottomrule
    \end{tabular}
    }
\end{table}

We also analyzed the running time of the MaxVol algorithm alone, including its chunk-based variant, under different initial numbers of frames, as shown in Table~\ref{tab:maxvol_runtime}. The experimental results show that the running time of MaxVol remains low across settings, with minimal impact on the overall inference efficiency.

\begin{table}[htb!]
    \centering
    \caption{MaxVol algorithm runtime (excluding image encoding time) for different input sizes.}
    \label{tab:maxvol_runtime}
    \small
    \resizebox{0.95\linewidth}{!}{
    \begin{tabular}{lcc}
        \toprule
        Method & \textbf{Input Size} & \textbf{MaxVol Time (s)} \\
        \midrule
        MaxInfo & 128 & 0.0044 \\
        MaxInfo & 256 & 0.0053 \\
        MaxInfo & 512 & 0.0109 \\
        Chunks-Based MaxInfo & $32 \times 32$ & 0.0375 \\
        \bottomrule
    \end{tabular}
    }
\end{table}

\begin{figure}[htb!]
    \centering
    \includegraphics[width=\linewidth]{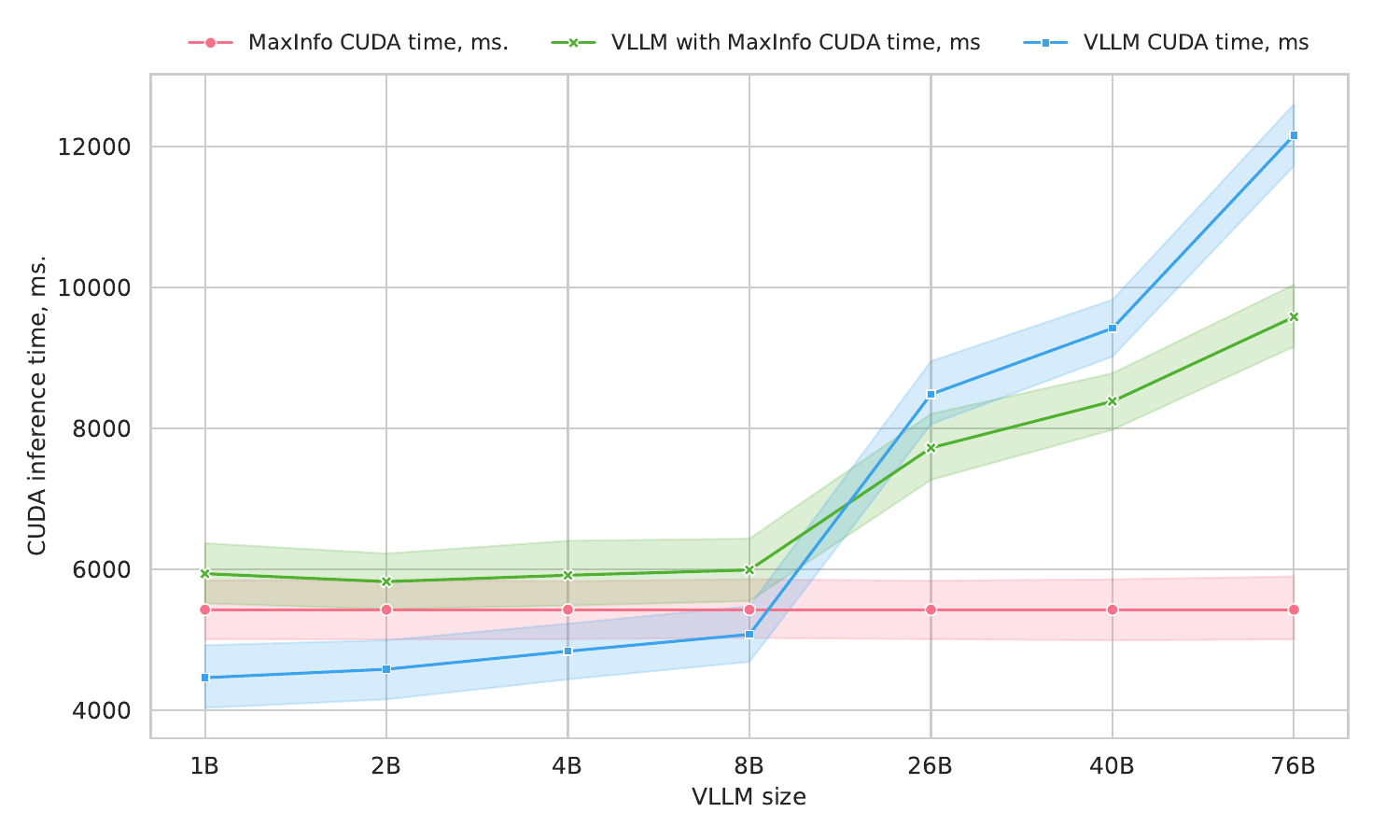}
    \caption{CUDA inference time across different VLLM sizes. The preprocessing cost of MaxInfo remains small and nearly constant, while overall inference time increases with model size.}
    \label{inference-time}
\end{figure}

Then we estimated CUDA inference time across different VLLM sizes which is shown in Figure~\ref{inference-time}. The overhead of MaxInfo remains small and nearly constant, while the overall inference time grows with model size, demonstrating that MaxInfo adds minimal cost compared to the savings from reduced visual tokens. For small models (up to 8B parameters), the relative benefit is limited since inference cost is low. However, for larger models (26B–76B), MaxInfo provides clear efficiency gains by substantially reducing the number of visual tokens, making its impact especially pronounced for long-video tasks where input length dominates computational cost.

\begin{figure}[htb!]
    \centering
    \includegraphics[width=\linewidth]{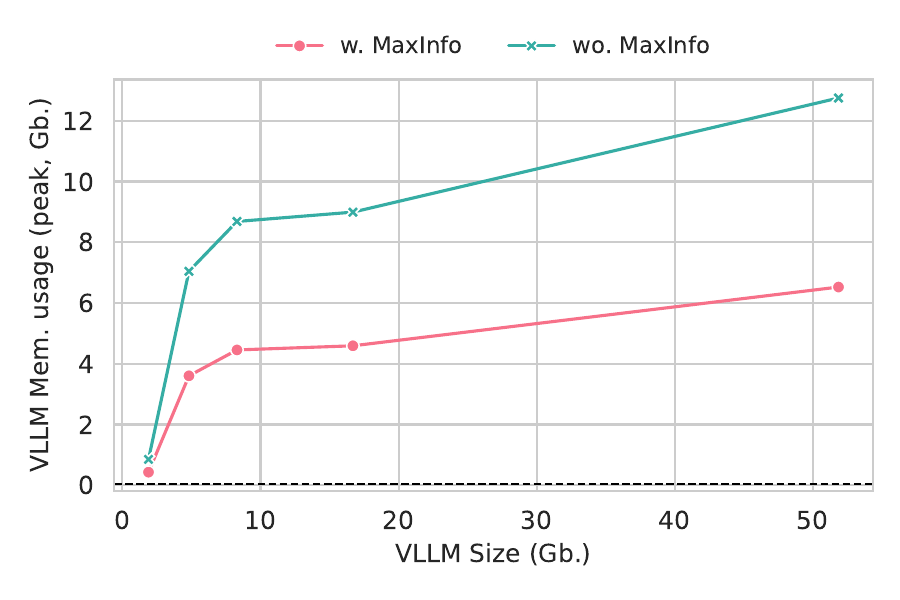}
    \caption{The comparison for memory performance on the GPU for InternVL2 models with and without MaxInfo module. The dashed line shows the CUDA memory requirements for MaxInfo.}
    \label{mem-estimation-fig}
\end{figure}

\textbf{ Memory Consumption.} Secondly, we precisely evaluated memory consumption of out approach. As shown in Figure~\ref{mem-estimation-fig}, MaxInfo’s CUDA memory usage remains constant for a fixed number of initial frames and grows much more slowly than uniform sampling as LLM size increases. 

In summary, our analysis of time and memory efficiency shows that MaxInfo introduces only negligible overhead while substantially reducing the computational burden of processing long videos. Its constant preprocessing cost and slower growth in memory usage make MaxInfo particularly advantageous for large-scale VLLMs, with the benefits becoming most pronounced for models exceeding 10B parameters.

\section{Theoretical Justification} \label{theoretical-justification}

\begin{definition}
Definition of the maximum volume of the video frame feature matrix.
\end{definition}

We consider a matrix $Q\in \mathbb{C}^{N \times r}$, where each row represents the CLIP or SigLIP etc. feature of a video frame, ordered sequentially in time, where N denotes the number of frames and r denotes the dimension of the feature.

We aim to identify a submatrix $\hat{Q} \in \mathbb{C}^{K \times r}$ of the original matrix $S$, such that $\hat{S}$ closely approximates $S$ in terms of matrix volume, thereby preserving its essential structural information.

To obtain the submatrix $\hat{Q}$, we introduce a coefficient matrix $C$ based on the minimum-norm linear combination as Equation~\ref{equations_a}
\begin{equation} \label{equations_a}
    \tilde{C} \hat{Q} = \tilde{Q}
\end{equation}
Here, $\tilde{Q}$ denotes a set of sample rows selected from the original matrix $Q$ for reconstruction. By solving for $\tilde{C}$, we can approximate the reconstruction of $\tilde{Q}$ using only the representative rows in $\hat{Q}$. In addition, it is shown that the selected K rows are the most representative of the video frame information.

\textbf{Solving.} The submatrix $\hat{Q} \in \mathbb{C}^{K \times r}$ provides an approximation of the original matrix $Q \in \mathbb{C}^{N \times r}$ within a tolerance $\tau$.

We start with an initial submatrix $\hat{Q} \in \mathbb{C}^{M \times r}$ and add a row $Q_i \in \mathbb{C}^{1 \times r}$ to each iteration to bring the expanded submatrix up to speed in the sense of volume. The updating process can be expressed as follows Equation~\ref{iter_M} and the volume of the updated matrix can be defined as Equation~\ref{new_vol}
\begin{equation} \label{iter_M}
    \hat{Q} \leftarrow
    \begin{bmatrix}
    \hat{Q} \\
    Q_i
    \end{bmatrix}
\end{equation}
\begin{equation} \label{new_vol}
    \text{Vol}(\hat Q)_{\text{new}} = \text{Vol}(\hat Q)_{\text{old}} \cdot \sqrt{1 + \| \tilde{C}_i \|_2^2}
\end{equation}
where $Q_i$ is the row selected from the original matrix $Q \in \mathbb{C}^{N \times r}$ that currently boosts the volume of the submatrix the most. Repeat this process iteratively until the conditional Equation~\ref{condition_iter} is satisfied or the target number of $K$ rows is reached.
\begin{equation}\label{condition_iter}
     \| \tilde{C}_i \|_2 \leq \tau
\end{equation}
\textbf{Proof of maximum information entropy.} To justify our approach, we use differential entropy as an information measure. Suppose our normalized frame embeddings form a matrix \( S \). The differential entropy of a uniform distribution over the convex hull \( \mathcal{C}(S) \) is given by the following Equation~\ref{TJ1}.
\begin{equation} \label{TJ1}
    H_{\max}(S) = \ln(\text{Vol}(\mathcal{C}(S)))
\end{equation}
where \( \text{Vol}(\mathcal{C}(S)) \) is the volume of the convex hull formed by selected embeddings. Classical results show the following Equation~\ref{TJ2}.
\begin{equation}\label{TJ2}
    \text{Vol}(\mathcal{C}(S)) = \kappa \sqrt{\det(S^\top S)}
\end{equation}
for some constant \( \kappa > 0 \). Thus we get Equation~\ref{TJ3}
\begin{equation}\label{TJ3}
    H_{\max}(S) = \ln V(S) + \text{constant}
\end{equation}
where \( V(S) = \sqrt{\det(S^\top S)} \). Since MaxVol maximizes \( V(S) \), it maximizes the upper bound on differential entropy, ensuring that selected frames are more informative.

In summary, we can theoretically select the most representative frame information. The feature matrices corresponding to the selected frames have good linear independence under the constraint of the tolerance parameter $\tau$, thus constituting an approximately optimal subset of the representation. This process achieves our goal of \textbf{information maximization}, i.e. preserving the most critical structural information while compressing redundancy.





\section{Societal Impacts} \label{society}

This work introduces a training-free framework for improving frame sampling in Vision-Language Large Models (VLLMs), enhancing video understanding tasks. Such advancements have important implications for applications in education, accessibility, and public safety.

However, improved video analysis capabilities may also raise ethical concerns, including potential misuse in surveillance, privacy violations, or biases affecting different communities. Ensuring responsible deployment with fairness and transparency is essential to mitigate these risks.

In summary, while our approach provides significant benefits, its adoption should adhere to ethical principles to promote equitable and responsible use.





\end{document}